\documentclass{article}

\PassOptionsToPackage{numbers,sort&compress}{natbib}
\usepackage[preprint]{neurips_2026}

\usepackage[utf8]{inputenc}
\usepackage[T1]{fontenc}
\usepackage[hidelinks]{hyperref}
\usepackage{url}
\usepackage{booktabs}
\usepackage{array}
\usepackage{graphicx}
\usepackage{amsmath}
\usepackage{multirow}
\usepackage{wrapfig}
\usepackage{placeins}
\usepackage{amsfonts}
\usepackage{nicefrac}
\usepackage{microtype}
\usepackage[table]{xcolor}
\usepackage{afterpage}
\usepackage{tikz}
\usepackage{listings}
\usepackage{float}
\usepackage{algorithm}
\usepackage{algpseudocode}
\usetikzlibrary{arrows.meta,positioning}

\newcommand{\bbcell}[1]{\raisebox{-0.5ex}{#1}}
\newcommand{\vbbcell}[2]{\shortstack[c]{ViT\\[-0.2pt]#1/#2}}
\definecolor{llmrow}{gray}{0.93}
\definecolor{qualgray}{RGB}{135,135,135}
\definecolor{qualred}{RGB}{218,31,31}
\definecolor{qualgreen}{RGB}{0,150,45}
\definecolor{qualblue}{RGB}{78,108,132}
\newcommand{\qualhead}[1]{{\rmfamily\bfseries\large #1}}
\newcommand{\qualvcenter}[1]{\raisebox{-.5\height}{#1}}
\newcommand{\qualmod}[1]{\qualvcenter{\rmfamily\small\shortstack[c]{#1}}}

\newcommand{\qualarrow}{\qualvcenter{\tikz[baseline=-0.5ex]{\draw[-{Stealth[length=5pt,width=6pt]},line width=1.6pt,qualblue] (0,0)--(0.38,0);}}}

\newcommand{\qualframeplain}[3]{%
  \begingroup
  \setlength{\fboxsep}{0pt}%
  \setlength{\fboxrule}{1.25pt}%
  \fcolorbox{#1}{white}{%
    \begin{minipage}[c][#2][c]{#2}
    \centering
    \includegraphics[width=\linewidth,height=\linewidth,keepaspectratio]{#3}%
    \end{minipage}}%
  \endgroup
}
\newcommand{\qualmodplain}[1]{\raisebox{-0.90em}{\rmfamily\small\shortstack[c]{#1}}}
\newcommand{\qualarrowplain}{\tikz[baseline=-0.5ex]{\draw[-{Stealth[length=5pt,width=6pt]},line width=1.6pt,qualblue] (0,0)--(0.38,0);}}
\newcommand{\fashioncrop}[4]{%
  \qualvcenter{\includegraphics[trim=#1bp #2bp #3bp #4bp,clip,width=0.135\textwidth]{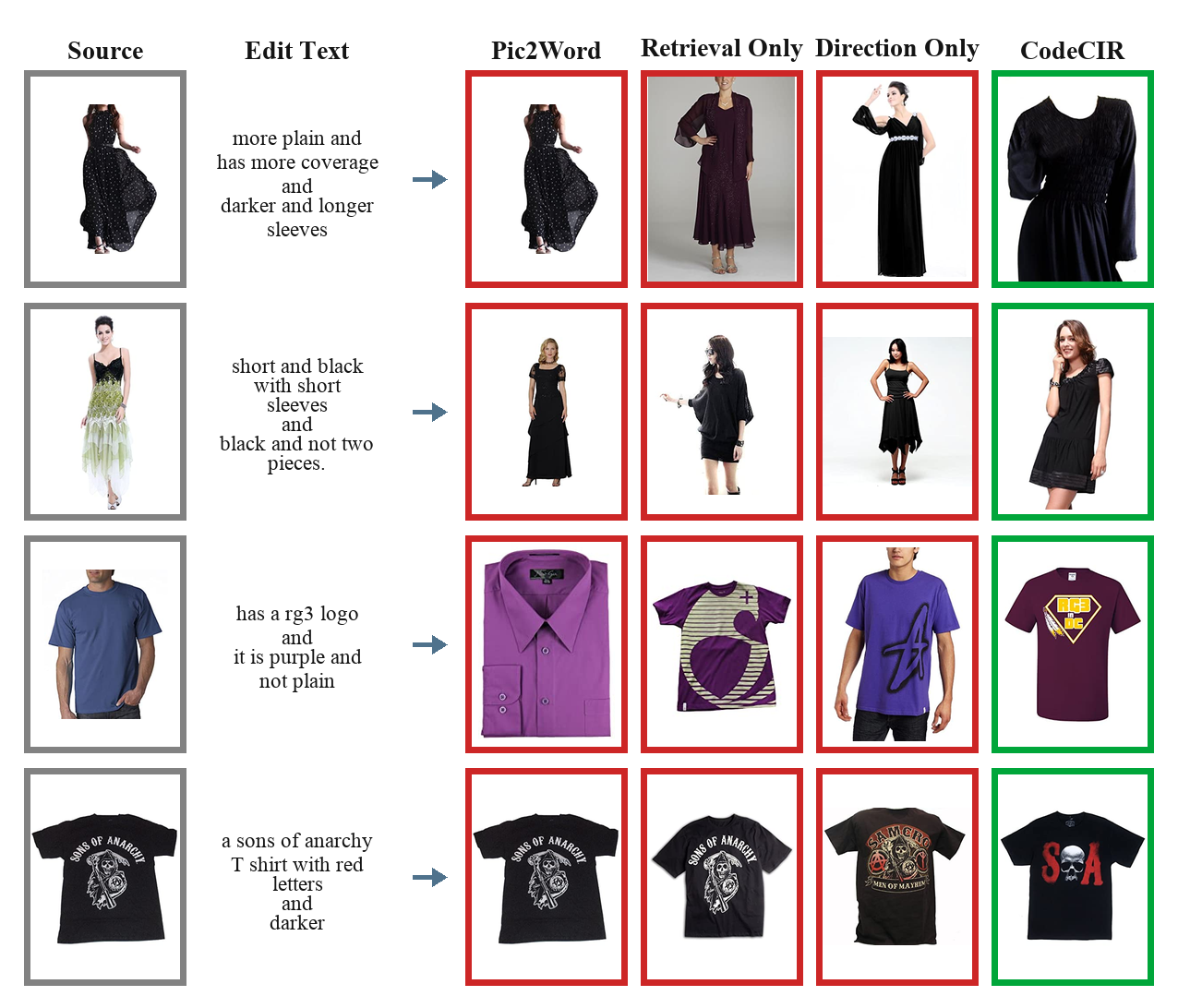}}%
}

\lstdefinestyle{promptstyle}{
  basicstyle=\footnotesize\rmfamily,
  breaklines=true,
  breakatwhitespace=false,
  columns=fullflexible,
  keepspaces=true,
  frame=single,
  framerule=0.75pt,
  framesep=4pt,
  rulecolor=\color{black!90},
  showstringspaces=false,
  xleftmargin=0pt,
  xrightmargin=0pt,
  aboveskip=0.35em,
  belowskip=0.35em
}
\newenvironment{contributionlist}{
  \begin{list}{$\bullet$}{
    \setlength{\leftmargin}{1.25em}
    \setlength{\labelwidth}{0.65em}
    \setlength{\labelsep}{0.45em}
    \setlength{\itemsep}{0.2em}
    \setlength{\topsep}{0.25em}
    \setlength{\parsep}{0pt}
    \setlength{\partopsep}{0pt}
  }
}{
  \end{list}
}

\title{Decoupling Endpoint and Semantic Transition Learning for Zero-Shot Composed Image Retrieval}

\author{%
{\normalsize\bfseries Mingyu Liu\textsuperscript{\bfseries 1} \quad Sihan Huang\textsuperscript{\bfseries 1} \quad Yijia Fan\textsuperscript{\bfseries 1} \quad
Yinlin Yan\textsuperscript{\bfseries 1} \quad Quan Zhang\textsuperscript{\bfseries 1}}\\[0.05em]
{\normalsize\bfseries Jian-Fang Hu\textsuperscript{\bfseries 1,2,3}\thanks{Corresponding author.} \quad Jianhuang Lai\textsuperscript{\bfseries 1,2,3}}\\[0.12em]
{\small\normalfont \textsuperscript{1}Sun Yat-sen University}\\[-0.02em]
{\small\normalfont \textsuperscript{2}Guangdong Province Key Laboratory of Information Security Technology, China}\\[-0.02em]
{\small\normalfont \textsuperscript{3}Key Laboratory of Machine Intelligence and Advanced Computing, Ministry of Education, China}
}

\begin{document}

\maketitle

\vspace{-0.6em}

\begin{abstract}
\begingroup\parfillskip=0pt\relax
Zero-shot composed image retrieval (ZS-CIR) retrieves a target image from a reference image and a text modification without human-annotated CIR triplets. Projection-based ZS-CIR methods are attractive because they do not rely on LLMs at inference and remain lightweight, but they often underperform LLM-based approaches on complex semantic modifications. This gap reflects a semantic transition bottleneck in projection-based ZS-CIR: endpoint-level matching can let the edit text act as a target-side attribute cue rather than grounding it as a source-conditioned semantic transition. We further show that adding semantic transition supervision to the same text adapter creates an endpoint--transition conflict between endpoint alignment and semantic transition alignment. To address this conflict, DeCIR decouples endpoint and transition learning. It constructs paired forward/reverse edit tuples from image-caption pairs, trains separate low-rank text adapter branches for endpoint alignment and semantic transition alignment, and merges them with Low-Rank Directional Merge (LRDM) into one deployable adapter. Extensive experiments on CIRR, CIRCO, FashionIQ, and GeneCIS demonstrate that DeCIR consistently improves projection-based ZS-CIR without increasing inference complexity. %Across CIRR, CIRCO, FashionIQ, and GeneCIS sets illustrate that DeCIR improves projection-based ZS-CIR without increasing inference complexity.
\par\endgroup
\end{abstract}

\vspace{-1.2em}

\section{Introduction}
\label{sec:introduction}
\begin{wrapfigure}[18]{r}{0.50\textwidth}
\vspace{-2.6em}
\centering
\includegraphics[width=\linewidth]{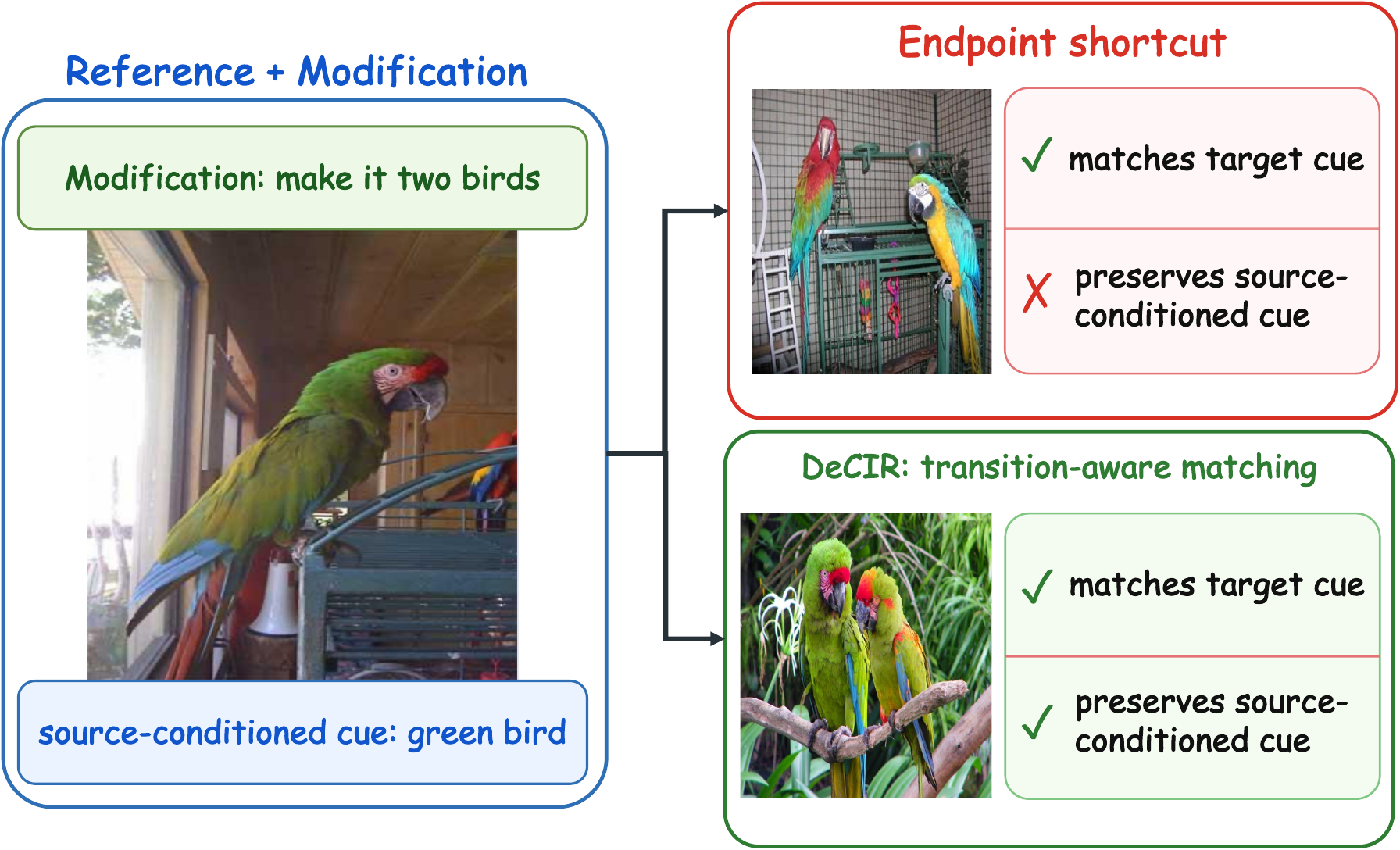}
{
\caption{\textbf{Illustration of our motivation.}
The semantic transition bottleneck in projection-based ZS-CIR can lead to an endpoint shortcut, where the edit text is treated as a target-side cue while source-conditioned evidence from the reference is dropped. This motivates DeCIR to explicitly learn transition-aware matching.}
}
\label{fig:intro-panel-a}
\vspace{-0.4em}
\end{wrapfigure}
Composed image retrieval (CIR) searches for a target image that matches both a reference image and a text modification. The target should apply the requested edit while preserving relevant visual evidence from the reference, i.e., it must be reachable from the reference under the instruction rather than merely satisfy an endpoint cue. Conventional CIR models learn this behavior from human-annotated reference--instruction--target triplets, which are expensive to collect and difficult to scale. Zero-shot composed image retrieval (ZS-CIR) instead trains on large image-caption datasets such as CC3M~\citep{cc3m}, but without triplets the model must infer source-to-target semantic transitions from weaker supervision. This makes the central supervision problem sharper: the model must infer how the source should change, even though the desired transition is not directly observed.

Existing ZS-CIR methods mainly follow two routes. LLM-based methods leverage the knowledge and reasoning abilities acquired during pretraining to infer a semantic transition from the source to the target~\citep{karthik2024vision,sun2025cotmr,huynh2025collm,tang2025reason,wang2026wiser}. By rewriting the composed query into a target caption, LLMs serve as a bridge from the reference--modification input to a hypothesized target endpoint, making them effective at handling complex edits, but often at the cost of heavier inference pipelines. Projection-based methods take a much lighter route: they map the reference image into the text embedding space of a pretrained vision-language model such as CLIP, combine it with the modification text, and encode the resulting prompt into a composed query embedding for retrieval~\citep{saito2023pic2word,baldrati2023zero,tang2024context,suo2024knowledge,zhong2025zero}. This route is attractive for deployment because it avoids heavy LLM calls and keeps the retrieval pipeline lightweight. However, by avoiding an explicit reasoning bridge, projection-based ZS-CIR leaves the source-to-target transition to be inferred implicitly by the CLIP text encoder, which encodes the pseudo-word and modification into a retrieval embedding rather than explicitly constructing target semantics as an LLM can. This raises a key question: can such a lightweight projection-based pipeline reliably capture source-to-target semantic transitions for complex modifications?

We argue that existing projection-based methods remain limited by a \emph{semantic transition bottleneck}: The CLIP text encoder is expected to infer the source-to-target semantic transition, although such transitions are neither explicitly supervised in contrastive pretraining nor directly optimized during pseudo-word mapping network training. Without source-conditioned semantic transition supervision, the modification text can collapse into a target-side attribute cue under complex edits: the model may capture the target attribute in the instruction without binding it to the source condition provided by the reference image. For example, as shown in Figure~\ref{fig:intro-panel-a}, the reference image supplies the source-conditioned cue, a green bird, while the instruction is only ``make it two birds''. A projection-based model may treat the text as the endpoint cue ``two birds'' and retrieve any image containing two birds, instead of learning the semantic transition from one green bird in the reference image to two green birds. This shortcut can work in easy cases but becomes brittle when hard distractors share the same target attribute without satisfying the intended source-conditioned semantic transition.

Explicit semantic transition supervision is a natural way to address this bottleneck. However, we further show that adding it to the same adapter introduces a new optimization challenge. Endpoint alignment pulls the composed query toward target endpoints, whereas semantic transition alignment asks instruction embeddings to represent source-to-target displacement. These objectives are complementary for CIR, but forcing them through the same low-rank text adapter can cause gradient interference, which is especially pronounced in deeper text transformer blocks and thus degrade retrieval performance. This evidence motivates conflict-aware endpoint–transition decoupling: the model should allow endpoint alignment and transition alignment to specialize during training while keeping a lightweight projection-based retrieval pipeline at inference.

We propose DeCIR, a conflict-aware endpoint–transition decoupling framework for lightweight ZS-CIR. We leverage an LLM to convert image-caption pairs into paired forward/reverse edit tuples, providing diverse semantic transition supervision without human-annotated CIR triplets. DeCIR then trains two low-rank text-adapter branches over the CLIP text encoder: an endpoint alignment branch that aligns the composed query embedding with the target-caption endpoint, and a transition branch for instruction-level semantic displacement. Finally, Low-Rank Directional Merge (LRDM) folds the semantic transition alignment branch into the endpoint alignment branch, producing a single deployable text adapter. The resulting model preserves the standard projection-based inference pipeline while obtaining stronger source-conditioned transition awareness.

Overall, our contributions are summarized as follows:
\vspace{-0.4em}
\begin{contributionlist}
\item We identify a semantic transition bottleneck in projection-based ZS-CIR, where edit text is matched as a target-side cue rather than grounded as a source-conditioned transition.
\item We show that endpoint and transition objectives interfere in the same low-rank adapter, motivating decoupled endpoint and transition learning.
\item We introduce LRDM, a shared-basis coefficient merge that folds transition sensitivity into the projection-based retrieval pipeline while retaining lightweight inference.
\item Experiments on CIRR, CIRCO, FashionIQ, and GeneCIS show that DeCIR improves projection-based ZS-CIR without LLMs or reasoning loops at inference.
\end{contributionlist}
\vspace{-0.4em}
\section{Related Work}

\subsection{Zero-Shot Composed Image Retrieval}

Composed image retrieval was first studied with human-annotated reference–instruction–target triplets~\citep{vo2019composing,chen2020image,wu2021fashion,liu2021image,baldrati2022effective}. ZS-CIR removes these triplets and mainly follows two paths. Projection-based methods map the reference image into the text space and retrieve with a single composed query~\citep{saito2023pic2word,baldrati2023zero,tang2024context,suo2024knowledge,zhong2025zero}; related CIR work further improves composition modules, diffusion proxies, hard negatives, and robustness settings~\citep{baldrati2023combiner,gu2023compodiff,kwak2025qure,psomas2024rscir,sun2026benchcir}. These methods strengthen different parts of the endpoint-matching pipeline, such as source pseudo-word inversion, context-dependent visual mapping, or instruction-aware distillation. However, the modification itself is not explicitly isolated and supervised as a source-conditioned semantic transition. LLM-based methods leverage LLMs or unified multimodal models (UMMs) to reason over the composed query, rewrite it into a modified caption, or even synthesize imagined target images to improve retrieval~\citep{karthik2024vision,sun2025cotmr,huynh2025collm}. These methods~\citep{tang2025reason,li2025imagine,wang2026wiser} have demonstrated strong reasoning ability and impressive qualitative results, but such gains often come at the cost of heavy inference pipelines.

DeCIR instead stays in the projection-based setting and is closest to DistillCIR~\citep{zhong2025zero} in that both use LLM-generated supervision during training rather than LLMs at inference. The difference is the role assigned to that supervision. DistillCIR uses LLM-generated signals to distill the instruction awareness of LLMs into projection-based models, while DeCIR uses generated forward/reverse edits to explicitly supervise source-conditioned semantic displacement and studies how this transition objective interacts with existing endpoint matching.

\subsection{Vision-Language Adaptation and Model Merging}

Modern CIR methods rely heavily on pretrained vision-language models. Beyond CLIP, large-scale image-text pretraining has been improved through noisy web supervision, caption bootstrapping, frozen-encoder bootstrapping, and sigmoid contrastive losses~\citep{jia2021align,li2022blip,li2023blip2,zhai2023siglip}. Lightweight tuning then specializes these models without full retraining, using adapters, prefix tuning, visual prompt tuning, or multimodal prompt learning~\citep{houlsby2019adapters,li2021prefixtuning,jia2022vpt,khattak2023maple}. These techniques provide strong reusable representations and low-cost adaptation, but they do not by themselves specify how a modification instruction should encode a source-conditioned semantic transition. DeCIR uses this lightweight adaptation setting as the base, then separates endpoint and transition learning.

Model merging combines multiple fine-tuned models into a single checkpoint while preserving useful capabilities. Prior work studies weight averaging, importance-weighted merging, task-vector arithmetic, interference-aware merging, robust ability transfer, and low-cost merging settings~\citep{wortsman2022model,matena2022merging,ilharco2023editingmodelstaskarithmetic,yadav2023ties,yu2024language,zeng2025parameter}. DeCIR uses merging for a different purpose: the endpoint alignment branch defines the inference backbone, while the transition branch contributes semantic transition sensitivity learned from paired forward/reverse supervision. In this paper, we propose LRDM to merge low-rank coefficients while keeping the retrieval-oriented output basis fixed.

\section{Method}
\label{sec:method}

\begin{figure*}[t]
\centering
\makebox[\textwidth][c]{\includegraphics[width=1.1\textwidth]{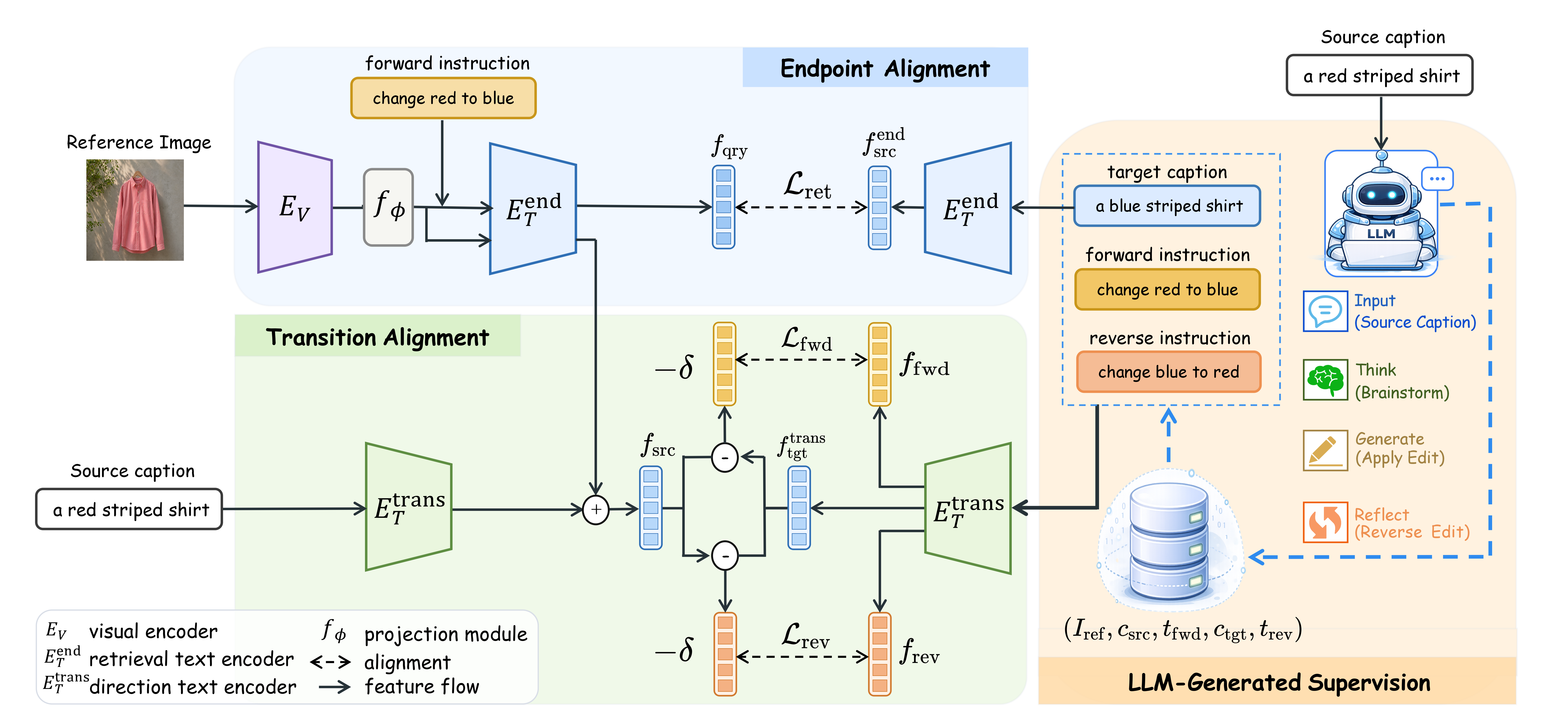}}
\caption{\textbf{Overview of DeCIR.} DeCIR decouples endpoint alignment and semantic transition alignment into two specialized branches during training. The text adapters in the two branches, i.e., $E_T^{\mathrm{end}}$ and $E_T^{\mathrm{trans}}$, are then merged into a single deployable adapter via Low-Rank Directional Merge, enabling efficient inference.
}
\label{fig:method_overview}
\end{figure*}

\subsection{Problem Formulation and Model Overview}
\label{sec:problem_overview}

Composed image retrieval aims to retrieve a target image from a reference image $I_{\mathrm{ref}}$ and a text modification $t$. DeCIR uses a projection-based CIR model based on the pretrained Pic2Word~\citep{saito2023pic2word}. Pic2Word uses a CLIP~\citep{radford2021learning} visual encoder $E_V$, a CLIP text encoder $E_T$, and a lightweight mapping network $f_{\phi}$ consisting of a three-layer MLP. Given $I_{\mathrm{ref}}$, the visual encoder first extracts an image feature, and the mapping network projects it into a pseudo-word token in the CLIP text embedding space, which is denoted as $s_{\mathrm{ref}}=f_{\phi}(E_V(I_{\mathrm{ref}}))$. Subsequently, this token is inserted into a text prompt with the modification, producing a composed query embedding. Retrieval is then performed by matching this query with cached gallery image embeddings. Unless otherwise noted, embeddings are $\ell_2$ normalized, and $S(a,b)=a^\top b$ denotes cosine similarity.

DeCIR keeps this lightweight projection-based inference form while adding endpoint alignment and semantic transition supervision during training. Instead of treating generated supervision only as a source-to-target transition, DeCIR constructs paired forward and reverse edits from each image-caption pair. The forward edit maps the source caption toward a synthetic target caption, while the reverse edit maps the target caption back toward the source semantics. The target caption supports endpoint alignment, and the forward/reverse instructions support semantic transition alignment. Since optimizing these two objectives in one shared text adapter can introduce interfering updates, DeCIR decouples endpoint alignment and semantic transition learning, then uses LRDM to fold them back into one deployable adapter, enabling efficient inference without any reliance on LLMs.

\subsection{Endpoint and Semantic Transition Alignment}
\label{sec:objectives}

A standard image-caption pair does not specify which semantic change should be applied to the source, so it cannot directly supervise source-conditioned semantic transitions. For each CC3M pair $(I_{\mathrm{ref}},c_{\mathrm{src}})$, we prompt an LLM~\citep{glm2024chatglm} to identify source details, brainstorm one plausible visual change, write a forward instruction, apply it to obtain a modified target caption, and write the reverse instruction that undoes only that change. This produces a bidirectional CIR-style tuple:
\begin{equation}
\left(I_{\mathrm{ref}},\; c_{\mathrm{src}},\; t_{\mathrm{fwd}},\; c_{\mathrm{tgt}},\; t_{\mathrm{rev}}\right).
\end{equation}
The target caption supplies the endpoint for endpoint alignment, while $t_{\mathrm{fwd}}$ and $t_{\mathrm{rev}}$ supply opposite semantic transitions. The tuple is bidirectional only at the instruction and text level: no reverse image retrieval pair is created, and the reverse instruction is used only for semantic transition supervision. Appendix~\ref{app:supervision_training} gives the detailed prompt and examples of llm-generated supervision.

\textbf{Endpoint alignment.}
The endpoint alignment branch learns to align the composed query with the generated target caption embedding, providing basic multimodal retrieval capability. Let $E_T^{\mathrm{end}}$ denote the endpoint alignment text encoder and $D_T$ the text embedding dimension. We use $\mathcal{T}_{\mathrm{cir}}(t)$ for a composed prompt such as ``a photo of * and $t$'' and $\mathcal{T}_{\mathrm{src}}(*)$ for the source-only prompt ``a photo of *,'' where $*$ is replaced by the pseudo token. For one training tuple, we define:
\begin{equation}
\begin{aligned}
f_{\mathrm{qry}}
&=E_T^{\mathrm{end}}\!\left(\mathcal{T}_{\mathrm{cir}}(t_{\mathrm{fwd}}),f_{\phi}(E_V(I_{\mathrm{ref}}))\right),\\
f_{\mathrm{tgt}}^{\mathrm{end}}
&=E_T^{\mathrm{end}}(c_{\mathrm{tgt}}),
\qquad f_{\mathrm{qry}}, f_{\mathrm{tgt}}^{\mathrm{end}}\in\mathbb{R}^{D_T}.
\end{aligned}
\end{equation}
Let $\mathcal{F}_{\mathrm{qry}}$ and $\mathcal{F}_{\mathrm{tgt}}^{\mathrm{end}}$ be the query and target-caption embeddings in the mini-batch. We optimize the symmetric contrastive loss
\begin{equation}
\mathcal{L}_{\mathrm{end}}
=
-\frac{1}{2}
\left[
\log
\frac{\exp(\tau S(f_{\mathrm{qry}}, f_{\mathrm{tgt}}^{\mathrm{end}}))}
{\sum_{\bar{f}\in\mathcal{F}_{\mathrm{tgt}}^{\mathrm{end}}}\exp(\tau S(f_{\mathrm{qry}}, \bar{f}))}
+
\log
\frac{\exp(\tau S(f_{\mathrm{tgt}}^{\mathrm{end}}, f_{\mathrm{qry}}))}
{\sum_{\bar{f}\in\mathcal{F}_{\mathrm{qry}}}\exp(\tau S(f_{\mathrm{tgt}}^{\mathrm{end}}, \bar{f}))}
\right],
\end{equation}
where $\tau$ is the learned temperature. This follows the common contrastive-learning form~\citep{chen2020simple,radford2021learning}, and the objective is averaged over the mini-batch.

\textbf{Transition alignment.}
The semantic transition alignment branch $E_T^{\mathrm{trans}}$ aims to teach edit instructions to encode the direction of change from the source semantics to the target semantics. We first compute the source and target anchors and use their difference as a transition proxy. The transition branch then learns to align forward and reverse instruction embeddings with this semantic displacement. The source anchor is built from the source caption and the reference-image-conditioned source prompt:
\begin{equation}
\begin{aligned}
f_{\mathrm{src}}^{\mathrm{cap}}
&=E_T^{\mathrm{trans}}(c_{\mathrm{src}}),
\quad f_{\mathrm{src}}^{\mathrm{cap}}\in\mathbb{R}^{D_T},\\
f_{\mathrm{src}}^{\mathrm{img}}
&=E_T^{\mathrm{trans}}\!\left(\mathcal{T}_{\mathrm{src}}(*),f_{\phi}(E_V(I_{\mathrm{ref}}))\right),
\quad f_{\mathrm{src}}^{\mathrm{img}}\in\mathbb{R}^{D_T},\\
f_{\mathrm{src}}
&=(1-\omega)f_{\mathrm{src}}^{\mathrm{cap}}+\omega f_{\mathrm{src}}^{\mathrm{img}},
\quad f_{\mathrm{src}}\in\mathbb{R}^{D_T}.
\end{aligned}
\end{equation}
Here $\omega$ controls how much image-conditioned evidence enters the anchor. Given this source anchor, the target caption provides the endpoint used to define the desired semantic transition. Notably, this endpoint is encoded by the semantic transition text encoder, and the source-to-target difference becomes the transition proxy:
\begin{equation}
\begin{aligned}
f_{\mathrm{tgt}}^{\mathrm{trans}}
&=E_T^{\mathrm{trans}}(c_{\mathrm{tgt}}),\quad
\delta=f_{\mathrm{tgt}}^{\mathrm{trans}}-f_{\mathrm{src}},
\quad f_{\mathrm{tgt}}^{\mathrm{trans}},\delta\in\mathbb{R}^{D_T}.
\end{aligned}
\end{equation}
Thus, $\delta$ is a source-to-target displacement for semantic transition. Since no target image is observed during CC3M training, this displacement is a text-embedding proxy for the desired visual edit. Before defining the transition loss, the semantic transition branch maps both generated edit instructions into the same text space:
\begin{equation}
\begin{aligned}
f_{\mathrm{fwd}}
&=E_T^{\mathrm{trans}}(t_{\mathrm{fwd}}),\quad
f_{\mathrm{rev}}
&=E_T^{\mathrm{trans}}(t_{\mathrm{rev}}),\quad
f_{\mathrm{fwd}},f_{\mathrm{rev}}\in\mathbb{R}^{D_T}.
\end{aligned}
\end{equation}
With these instruction embeddings, the forward edit is aligned with $\delta$ and the reverse edit is aligned with $-\delta$. The two transition losses are
\begin{equation}
\mathcal{L}_{\mathrm{fwd}}=1-S(f_{\mathrm{fwd}},\delta),
\qquad
\mathcal{L}_{\mathrm{rev}}=1-S(f_{\mathrm{rev}},-\delta).
\end{equation}
Here, the forward term makes the generated instruction follow the source-to-target displacement, while the reverse term uses the paired reverse instruction as an opposite-direction constraint. We combine the two terms to train the semantic transition branch for semantic transition alignment:
\begin{equation}
\mathcal{L}_{\mathrm{trans}}
=\mathcal{L}_{\mathrm{fwd}}+\mathcal{L}_{\mathrm{rev}}.
\label{eq:transition_loss}
\end{equation}
Together, the two terms encourage the modification text to encode a semantic displacement instead of serving only as a target-side attribute cue.

\subsection{Shared Adapter Interference}
\label{sec:interference}
A shared adapter can optimize $\mathcal{L}_{\mathrm{end}}+\mathcal{L}_{\mathrm{trans}}$ directly, but the two losses impose different local updates on the same low-rank factors. This can lead to gradient conflict: endpoint alignment and transition alignment may push the shared adapter in divergent directions, degrading retrieval performance. In this section, we diagnose this effect with an offline layer-wise gradient probe~\citep{hao2026unix} rather than relying on raw gradient similarity alone.

For an adapted text block $\ell$, write the LoRA update as $\Delta W^{\ell}=B^{\ell}A^{\ell}$. Starting from the same shared-adapter checkpoint, we aggregate and vectorize the gradients induced by $\mathcal{L}_{\mathrm{end}}$ and $\mathcal{L}_{\mathrm{trans}}$ over multiple mini-batches, denoted as $\bar{g}_{\mathrm{end}}^{\ell}$ and $\bar{g}_{\mathrm{trans}}^{\ell}$. Their cosine similarity measures the raw agreement between the endpoint and transition objectives. However, raw agreement can be biased by layer-dependent gradient stability, since different transformer blocks may naturally exhibit different levels of stochastic gradient consistency. To control for this effect, we also estimate a same-objective baseline by splitting gradients from the same objective into two disjoint groups, $\bar{g}_{A}^{\ell}$ and $\bar{g}_{B}^{\ell}$, where the subscripts denote the two gradient groups rather than the LoRA factors. The cross-objective and same-objective agreements are
\begin{equation}
s_{\mathrm{cross}}^{\ell}=S\!\left(\bar{g}_{\mathrm{end}}^{\ell},\bar{g}_{\mathrm{trans}}^{\ell}\right),
\qquad
s_{\mathrm{base}}^{\ell}=S\!\left(\bar{g}_{A}^{\ell},\bar{g}_{B}^{\ell}\right).
\end{equation}

\begin{wrapfigure}[12]{r}{0.48\textwidth}
\vspace{-0.3em}
\centering
\includegraphics[width=0.92\linewidth]{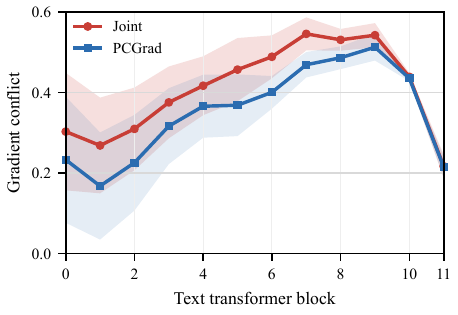}
\caption{\textbf{Layer-wise gradient interference}. Result is shown with mean $\pm$ std over five seeds.}
\label{fig:gradient_interference_probe}
\vspace{-0.8em}
\end{wrapfigure}
A larger interference score $\mathrm{GI}_{\ell}=s_{\mathrm{base}}^{\ell}-s_{\mathrm{cross}}^{\ell}$ means that endpoint and transition gradients agree less than expected under ordinary same-objective stochastic variation. In other words, a high $\mathrm{GI}_{\ell}$ indicates that the disagreement is not merely due to noisy gradients in that layer, but reflects additional conflict between the endpoint and transition objectives. Figure~\ref{fig:gradient_interference_probe} shows a clear conflict peak in middle and deep layers; PCGrad reduces the mean level but retains the same peak pattern. Together with the joint-training ablation in Table~\ref{tab:core_ablation}, this motivates separating the objectives during training and merging them only after specialization.
\subsection{Low-Rank Directional Merge}
\label{sec:decoupled_training}
\label{sec:lrdm}

DeCIR trains two text adapter branches initialized from the same CLIP text encoder. The endpoint alignment branch $E_T^{\mathrm{end}}$ optimizes $\mathcal{L}_{\mathrm{end}}$ and defines the deployable composed retrieval pathway, while the transition branch $E_T^{\mathrm{trans}}$ optimizes $\mathcal{L}_{\mathrm{trans}}$ and learns semantic instruction displacement. The visual encoder LoRA parameters and the inversion network $f_{\phi}$ belong to the retrieval pathway and are updated only by $\mathcal{L}_{\mathrm{end}}$.

LRDM makes the later merge well defined by sharing the low-rank output basis. For each adapted text transformer layer $\ell$, the two branches share $B^{\ell}$ but keep branch-specific coefficients:
\begin{equation}
\Delta W_{\mathrm{end}}^{\ell}=B^{\ell}A_{\mathrm{end}}^{\ell},
\qquad
\Delta W_{\mathrm{trans}}^{\ell}=B^{\ell}A_{\mathrm{trans}}^{\ell}.
\end{equation}
Training uses two sequential passes over the same mini-batch. The endpoint alignment branch updates $\{A_{\mathrm{end}}^{\ell}\}$, $\{B^{\ell}\}$, the visual LoRA parameters, $f_{\phi}$, and $\tau$; the transition alignment branch updates only $\{A_{\mathrm{trans}}^\ell\}$. Thus, transition supervision changes semantic transition coefficients without rewriting the retrieval basis or visual pathway.

After training, LRDM restores a single low-rank adapter by merging coefficients rather than full weights:
\begin{equation}
A_{\mathrm{merge}}^{\ell}
=(1-\alpha)A_{\mathrm{end}}^{\ell}+\alpha A_{\mathrm{trans}}^{\ell}.
\end{equation}
The deployable layer keeps the shared basis and absorbs the merged update:
\begin{equation}
B_{\mathrm{merge}}^{\ell}=B^{\ell},\qquad
W'^{\ell}=W^{\ell}+B_{\mathrm{merge}}^{\ell}A_{\mathrm{merge}}^{\ell}.
\end{equation}
Here $\alpha$ controls how much semantic transition information is injected into the low-rank adapter. LRDM is asymmetric: the endpoint alignment branch keeps the backbone and shared basis, while the transition branch contributes semantic transition coordinates inside that basis. The transition branch is then discarded; gallery embeddings are encoded once with the final visual encoder and cached, and query retrieval uses only the merged pathway. Algorithm~\ref{alg:decir} in Appendix~\ref{app:supervision_training} summarizes the full training and merge procedure.

\section{Experiments}

\begin{table}[!t]
\centering
\caption{\textbf{Quantitative comparison on the test set of CIRCO and the validation set of Fashion-IQ.} We report the mAP@$K$ metrics for CIRCO test set and the R@$K$ metrics for Fashion-IQ validation set across ViT-B/32 and ViT-L/14 backbones. Best results are in bold, second-best results are underlined, and our method is highlighted in grey.}
\footnotesize
\renewcommand{\arraystretch}{1.20}
\setlength{\tabcolsep}{1.9pt}
\resizebox{\textwidth}{!}{
\begin{tabular}{ll*{4}{c}*{8}{c}}
\toprule
\multirow{3}{*}{} & \multirow{3}{*}{\hspace{-2pt}Method} & \multicolumn{4}{c}{CIRCO} & \multicolumn{8}{c}{FashionIQ} \\
\cmidrule(lr){3-6} \cmidrule(lr){7-14}
 &  & \multicolumn{4}{c}{mAP@k} & \multicolumn{2}{c}{Dress} & \multicolumn{2}{c}{Shirt} & \multicolumn{2}{c}{TopTee} & \multicolumn{2}{c}{Avg.} \\
 & \multicolumn{1}{l}{} & k=5 & k=10 & k=25 & k=50 & R@10 & R@50 & R@10 & R@50 & R@10 & R@50 & R@10 & R@50 \\
\midrule
\multirow[c]{4}{*}{\bbcell{ViT-B/32}}
& SEARLE (ICCV'23) & 9.4 & 9.9 & 11.1 & 11.8 & 18.5 & 39.5 & 24.4 & 41.6 & 25.7 & 46.5 & 22.9 & 42.5 \\
& Slerp (ECCV'24) & 9.3 & 10.3 & 11.7 & 12.3 & 19.2 & 42.1 & 23.0 & 42.0 & 26.6 & 47.8 & 23.0 & 44.0 \\
& CIReVL (ICLR'24) & 14.9 & 15.4 & 17.0 & 17.8 & \textbf{25.3} & \textbf{46.4} & 28.4 & 47.8 & 31.2 & \underline{53.9} & 28.3 & \underline{49.4} \\
& DistillCIR (ICCV'25) & \underline{16.3} & \underline{16.9} & \underline{18.5} & \underline{19.4} & 25.1 & 45.8 & \underline{29.3} & \underline{49.1} & \underline{33.9} & 53.1 & \underline{29.5} & 49.3 \\
\rowcolor{llmrow}& \textbf{DeCIR (ours)} & \textbf{17.1} & \textbf{17.6} & \textbf{19.1} & \textbf{20.0} & \underline{25.2} & \underline{46.0} & \textbf{29.9} & \textbf{49.4} & \textbf{33.9} & \textbf{54.1} & \textbf{29.7} & \textbf{49.8} \\
\midrule
\multirow[c]{9}{*}{\bbcell{ViT-L/14}}
& Pic2Word (CVPR'23) & 8.7 & 9.5 & 10.6 & 11.3 & 20.0 & 40.2 & 26.2 & 43.6 & 27.9 & 47.4 & 24.7 & 43.7 \\
& SEARLE (ICCV'23) & 11.7 & 12.7 & 14.3 & 15.1 & 20.5 & 43.2 & 27.4 & 45.7 & 29.3 & 50.2 & 25.7 & 46.4 \\
& LinCIR (CVPR'24) & 12.6 & 13.6 & 15.0 & 15.9 & 20.9 & 42.4 & 29.1 & 46.8 & 28.8 & 50.2 & 26.3 & 46.5 \\
& Context-I2W (AAAI'24) & 13.0 & 14.6 & 16.1 & 17.2 & 23.1 & 45.3 & 29.7 & 48.6 & 30.6 & 52.9 & 27.8 & 48.9 \\
& Slerp (ECCV'24) & 18.5 & 19.4 & 21.4 & 22.4 & 23.4 & 45.1 & 29.6 & 46.5 & 32.0 & 51.2 & 28.3 & 47.6 \\
& CIReVL (ICLR'24) & 18.6 & 19.0 & 20.9 & 21.8 & 24.8 & 44.8 & 29.5 & 47.4 & 31.4 & 53.7 & 28.5 & 48.6 \\
& MOA (SIGIR'25) & 15.3 & 17.1 & 18.5 & 19.3 & 25.2 & \underline{48.5} & \underline{31.9} & \underline{50.7} & 33.2 & \underline{54.8} & 30.1 & \underline{51.3} \\
& HIT (ICCV'25) & 15.5 & 16.7 & 18.9 & 19.9 & \underline{25.6} & 47.1 & \textbf{32.4} & \textbf{51.2} & 32.8 & 54.7 & \underline{30.3} & 51.0 \\
& DistillCIR (ICCV'25) & \underline{18.7} & \underline{20.3} & \underline{22.8} & \underline{23.9} & 25.5 & 46.6 & 30.7 & 50.1 & \underline{33.9} & 53.7 & 30.0 & 50.2 \\
\rowcolor{llmrow}& \textbf{DeCIR (ours)} & \textbf{21.3} & \textbf{22.5} & \textbf{24.7} & \textbf{25.7} & \textbf{25.8} & \textbf{49.8} & 31.2 & 50.2 & \textbf{34.1} & \textbf{54.9} & \textbf{30.4} & \textbf{51.6} \\
\bottomrule
\end{tabular}
}

\label{tab:circo_fashioniq_main}
\vspace{-0.8em}
\end{table}
\renewcommand{\arraystretch}{1.0}

\subsection{Experimental Setup}

For data processing, we prompt the lightweight LLM \texttt{glm-4-flash} to generate a forward edit, a modified target caption, and a reverse edit as described in Section~\ref{sec:objectives}, obtaining 667,229 clean tuples after format filtering. To ensure supervision quality, we manually review random samples for edit fidelity and reverse-instruction consistency. Appendix~\ref{app:supervision_training} provides the prompt, examples, and other details. We evaluate DeCIR with CLIP ViT-B/32 and ViT-L/14 backbones. Following common projection-based baselines, we train on CC3M and initialize from the pretrained Pic2Word~\citep{saito2023pic2word} checkpoint and employ AdamW \citep{loshchilov2018decoupled} with a learning rate of $2 \times 10^{-5}$, weight decay of 0.1, and a linear warmup of 200 steps. The batch size is 768. The training is conducted on 4 NVIDIA A100 (80G) GPUs for one epoch with cosine learning rate scheduling. For parameter-efficient training, we train the full Pic2Word mapping network while applying LoRA~\citep{hu2022lora} to both visual and text encoder projection layers in CLIP. We report the performance averaged over three trials.

\subsection{Benchmarks and Baselines}

We evaluate DeCIR on four CIR benchmarks that cover different aspects of composed retrieval: CIRR for natural image disambiguation, CIRCO for multi-target ranking, FashionIQ for fashion edits, and GeneCIS for conditional retrieval. Following established practice, we report Recall@k (R@k) for CIRR, FashionIQ, and GeneCIS, with the CIRR subset metric (R$_s$@k) measuring retrieval within a curated candidate set. Because CIRCO can include multiple correct targets for a single query, we report mean Average Precision (mAP@k). We compare DeCIR with representative projection-based methods, including Pic2Word~\citep{saito2023pic2word}, SEARLE~\citep{baldrati2023zero}, LinCIR~\citep{gu2024language}, Context-I2W~\citep{tang2024context}, Slerp~\citep{slerp}, HIT~\citep{li2025hierarchy}, MOA~\citep{moa}, PrediCIR~\citep{predicir}, and DistillCIR~\citep{zhong2025zero}, as well as the LLM-based CIReVL~\citep{karthik2024vision}. Due to space limitations, additional results are provided in Appendix~\ref{app:additional_benchmark_results}.

\par\noindent\textbf{CIRCO.}\quad
CIRCO includes multiple valid targets per query, making ranking quality particularly important for composed retrieval. DeCIR consistently achieves the best mAP across all evaluation cutoffs with both ViT-B/32 and ViT-L/14 backbones. With ViT-B/32, DeCIR improves over the strongest prior method by 0.8\%, 0.7\%, 0.6\%, and 0.6\% on mAP@5, mAP@10, mAP@25, and mAP@50, respectively. The gains become more pronounced with ViT-L/14, where DeCIR achieves 21.3\% mAP@5 and improves over DistillCIR by 2.6\%. Across all cutoffs, DeCIR outperforms the best competing method by 2.6\%, 2.2\%, 1.9\%, and 1.8\% on mAP@5, mAP@10, mAP@25, and mAP@50, respectively. These consistent improvements indicate that the proposed transition modeling is particularly effective for open-domain composed retrieval, where the target is determined by fine-grained semantic changes rather than category-level similarity alone.

\par\noindent\textbf{FashionIQ.}\quad
FashionIQ focuses on fine-grained attribute modifications within the fashion domain, covering categories such as dresses, shirts, and toptees. DeCIR obtains competitive or superior performance across the three fashion categories. With ViT-B/32, DeCIR achieves the best average results, reaching 29.7\% R@10 and 49.8\% R@50, while also producing clear gains on Shirt and TopTee. With ViT-L/14, DeCIR further improves the average performance to 30.4\% R@10 and 51.6\% R@50, achieving the best average performance among all compared methods. These results suggest that the learned semantic transition transfers beyond the natural-image CC3M training distribution and provides a more robust training signal for zero-shot composed retrieval in fine-grained scenarios.

\newpage
\begin{wraptable}[22]{r}{0.55\textwidth}
\vspace{-1.4em}
\centering
\caption{\textbf{Quantitative comparison on the CIRR test set.} We compare our DeCIR with state-of-the-art methods across ViT-B/32 and ViT-L/14 backbones. Best results are in bold, second-best results are underlined, and our method is highlighted in grey.}
\scriptsize
\renewcommand{\arraystretch}{0.90}
\setlength{\tabcolsep}{1.10pt}
\resizebox{\linewidth}{!}{%
\begin{tabular}{@{}c@{\hspace{2pt}}l*{3}{c}*{3}{c}@{}}
\toprule
\multirow{2}{*}{} & \multirow{2}{*}{Method} & \multicolumn{3}{c}{Recall@k} & \multicolumn{3}{c}{RecallSubset@k} \\
\cmidrule(lr){3-5} \cmidrule(lr){6-8}
 &  & R@1 & R@5 & R@10 & R$_s$@1 & R$_s$@2 & R$_s$@3 \\
\midrule
\multirow[c]{5}{*}{\vbbcell{B}{32}} & SEARLE & 24.0 & 53.4 & 66.8 & 54.9 & 76.6 & 88.2 \\
& Slerp & 28.2 & 55.9 & 68.8 & 61.1 & 80.6 & 90.7 \\
 & CIReVL & 23.9 & 52.5 & 66.0 & 60.2 & 80.1 & 90.2 \\
 & DistillCIR & \underline{29.9} &\underline{58.8} & \underline{70.3} & \underline{62.3}& \underline{82.5} & \underline{90.8}\\
\rowcolor{llmrow} & \textbf{DeCIR (ours)} & \textbf{30.7} & \textbf{59.8} & \textbf{72.4} & \textbf{63.1} & \textbf{83.3} & \textbf{91.3} \\
\midrule
\multirow[c]{10}{*}{\vbbcell{L}{14}} & Pic2Word & 23.9 & 51.7 & 65.3 & 53.8 & 74.5 & 87.1 \\
 & SEARLE & 24.2 & 52.5 & 66.3 & 53.8 & 75.0 & 88.2 \\
 & LinCIR & 25.0 & 53.3 & 66.7 & 57.1 & 77.4 & 88.9 \\
 & Context-I2W & 25.6 & 55.1 & 68.5 & -& - &- \\
 & Slerp & 30.9 & 59.4 & 70.9 & 64.7 & 82.9 & \textbf{92.3} \\
 & CIReVL & 24.6 & 52.3 & 64.9 & 59.5 & 79.9 & 89.7 \\
 & MOA & 27.1 & 56.5 & 69.2 & - & - & - \\
 & HIT & 27.9 & 57.6 & 70.5 & - & - & - \\
 & DistillCIR & \underline{32.3} &\underline{63.6} & \underline{74.3} & \underline{66.8} & \underline{84.0} & 91.8\\
\rowcolor{llmrow} & \textbf{DeCIR (ours)} & \textbf{33.4} & \textbf{64.4} & \textbf{74.8} & \textbf{67.1} & \textbf{84.7} & \underline{92.2}\\
\bottomrule
\end{tabular}
}
\setlength{\abovecaptionskip}{3pt}

\label{tab:cirr_main}
\vspace{-0.9em}
\end{wraptable}
\renewcommand{\arraystretch}{1.0}

\noindent\textbf{CIRR.}\quad
CIRR contains real-world natural scenes where the instruction must disambiguate the target from visually related candidates. As shown in Table~\ref{tab:cirr_main}, DeCIR reaches 33.4\% R@1 with ViT-L/14. On the subset protocol, where the target is retrieved from six curated candidates, DeCIR achieves 67.1\% R$_s$@1. These results suggest that explicit semantic transition alignment is helpful when retrieval depends on fine-grained edits rather than coarse visual similarity alone.

\FloatBarrier

\subsection{Ablation Study}
We evaluate the contribution of the core components in DeCIR. We first test whether endpoint alignment and transition alignment should be trained jointly or decoupled. We then compare merge and adaptation variants. Unless otherwise specified, we use ViT-L/14 as the backbone and report validation set results. Additional ablations and analyzes are provided in Appendix~\ref{app:additional_ablations}.

\paragraph{Effect of Decoupled Training}

To understand the contribution of each training branch, we compare five settings: transition-only training with $\mathcal{L}_{\mathrm{trans}}$, retrieval-only training with $\mathcal{L}_{\mathrm{end}}$, joint training in one shared adapter, joint training with PCGrad~\citep{yu2020gradient}, and the full decoupled approach followed by LRDM. All variants in Table~\ref{tab:core_ablation} use the same backbone and paired supervision source; the PCGrad row applies gradient projection to the shared-adapter joint objective.

\begin{table}[H]
\centering
\caption{\textbf{Ablation study on decoupling endpoint and semantic transition learning.} All variants are trained with the same backbone and data, differing only in the optimization and merging strategy for endpoint alignment and semantic transition alignment.}
\scriptsize
\renewcommand{\arraystretch}{1.10}
\setlength{\tabcolsep}{5pt}
\begin{tabular}{lcc l ccc}
\toprule
Method & $\mathcal{L}_{\mathrm{end}}$ & $\mathcal{L}_{\mathrm{trans}}$ & Inference & CIRR & CIRCO & GeneCIS \\
\cmidrule(lr){5-7}
 & & & & R$_s$@1 & mAP@50 & Avg. R@1 \\
\midrule
Transition only & $\times$ & $\checkmark$ & transition branch & 54.92 & 13.27 & 13.20 \\
Endpoint only & $\checkmark$ & $\times$ & endpoint branch & 64.46 & 24.92 & 16.50 \\
Joint training & $\checkmark$ & $\checkmark$ & shared adapter & 61.09 & 21.93 & 16.27 \\
Joint training (PCGrad) & $\checkmark$ & $\checkmark$ & shared adapter & 64.20 & 24.30 & 16.43 \\
Decoupled + LRDM & $\checkmark$ & $\checkmark$ & merged adapter & 66.08 & 27.26 & 16.86 \\
\bottomrule
\end{tabular}

\label{tab:core_ablation}
\end{table}

Table~\ref{tab:core_ablation} reveals three distinct patterns that validate our architectural design. First, transition-only training performs substantially worse than retrieval-only training. This is theoretically expected, as $\mathcal{L}_{\mathrm{trans}}$ is designed to force instruction embeddings to capture relative semantic displacements rather than optimizing full composed queries to locate absolute gallery endpoints. Thus, semantic transition learning acts as a complementary signal to enrich endpoint alignment, rather than a standalone retrieval objective. Second, naively combining these objectives highlights a critical optimization bottleneck: shared-adapter joint training actively underperforms the retrieval-only baseline. This degradation indicates severe gradient interference when semantic transition supervision and endpoint alignment compete for the same low-rank parameter space. While applying gradient surgery (PCGrad)~\citep{yu2020gradient} partially mitigates this conflict on CIRR and GeneCIS, it remains far behind the retrieval-only baseline on CIRCO. Third, our proposed strategy—decoupled training followed by LRDM—achieves the best results on CIRR and CIRCO while maintaining the strongest overall balance. This conclusively demonstrates that semantic transition supervision is most effective when isolated to allow branch specialization during training and systematically integrated only after convergence.

\newpage
\paragraph{Effect of LRDM}

\begin{wraptable}[15]{r}{0.515\textwidth}
\vspace{-1.4em}
\centering
\caption{\textbf{Merge strategy ablation.} Branches are fixed; only the merge rule changes. All merge weights are set to 0.5; Task Arithmetic is the plain weight-space merge baseline.}
\small
\renewcommand{\arraystretch}{1.08}
\setlength{\tabcolsep}{1.2pt}
\begin{tabular}{@{}lcc@{}}
\toprule
Merge strategy & CIRR R$_s$@1 & CIRCO mAP@50 \\
\midrule
Task Arithmetic~\citep{ilharco2023editingmodelstaskarithmetic} & 65.22 & 27.07 \\
TIES~\citep{yadav2023ties} & 63.29 & 26.70 \\
DARE~\citep{yu2024language} & 64.96 & 27.05 \\
DARE-TIES~\citep{yu2024language,yadav2023ties} & 63.38 & 26.75 \\
Model Breadcrumbs~\citep{modelbreadcrumbs} & 64.41 & 27.14 \\
RobustMerge~\citep{zeng2025parameter} & 65.20 & 27.06 \\
LRDM & 66.08 & 27.26 \\
\bottomrule
\end{tabular}

\label{tab:merge_main}
\vspace{-0.4em}
\end{wraptable}

LRDM integrates the transition branch into a single deployable low-rank adapter. Table~\ref{tab:merge_main} compares LRDM with representative generic merge rules while keeping the trained branches fixed. LRDM performs best on both validation metrics, supporting the asymmetric merge design: the endpoint alignment branch preserves the endpoint-aligned backbone, while the transition branch injects semantic transition coefficients in the shared low-rank basis. Thus, the gain does not come from arbitrary model averaging. Appendix~\ref{app:additional_ablations} reports the full merge table with additional rules and metrics, plus merge-weight sensitivity and trainable component scope.

\begin{figure}[H]
\centering
\setlength{\tabcolsep}{3.4pt}
\renewcommand{\arraystretch}{0.92}
\begin{tabular}{@{}>{\centering\arraybackslash}m{0.155\textwidth}
                  >{\centering\arraybackslash}m{0.205\textwidth}
                  >{\centering\arraybackslash}m{0.030\textwidth}
                  >{\centering\arraybackslash}m{0.155\textwidth}
                  >{\centering\arraybackslash}m{0.155\textwidth}@{}}
\multicolumn{1}{c}{\qualhead{Reference}} &
\multicolumn{1}{c}{\qualhead{Modification}} & &
\multicolumn{1}{c}{\qualhead{Pic2Word}} &
\multicolumn{1}{c}{\qualhead{DeCIR}} \\
\qualframeplain{qualgray}{0.135\textwidth}{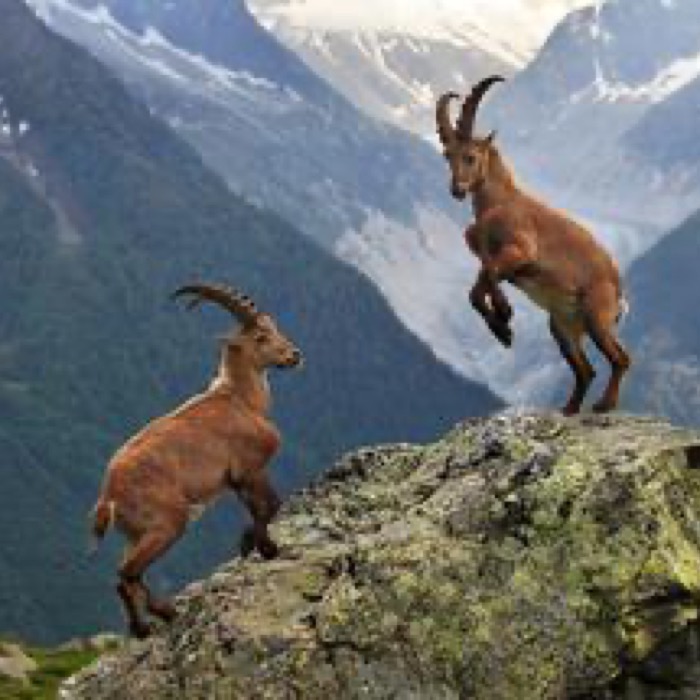} &
\qualmodplain{change to a herd\\of three wild\\mountain goats\\with a tighter rocky\\mountain view} &
\qualarrowplain &
\qualframeplain{qualred}{0.135\textwidth}{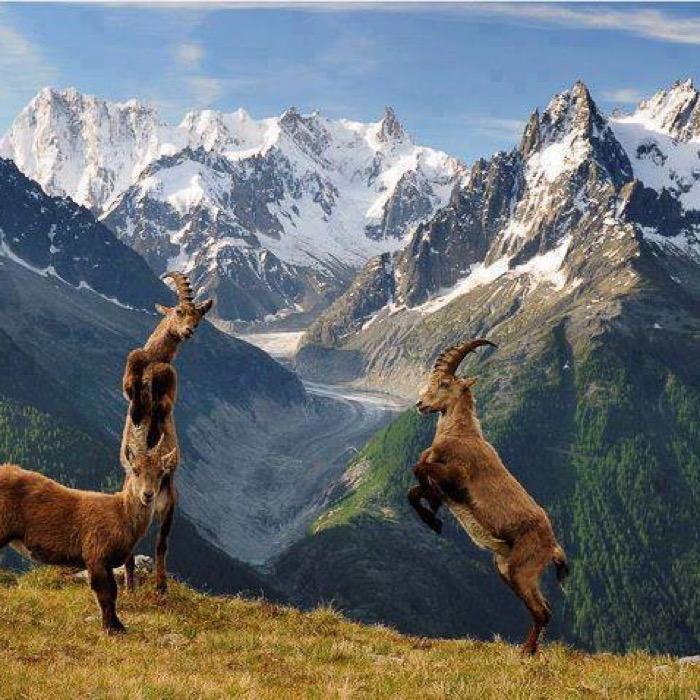} &
\qualframeplain{qualgreen}{0.135\textwidth}{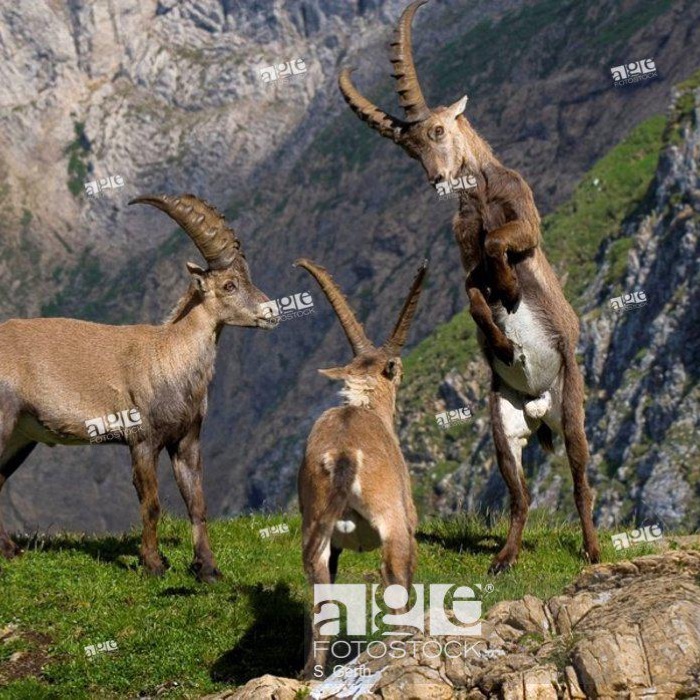} \\
\addlinespace[0.30em]
\qualframeplain{qualgray}{0.135\textwidth}{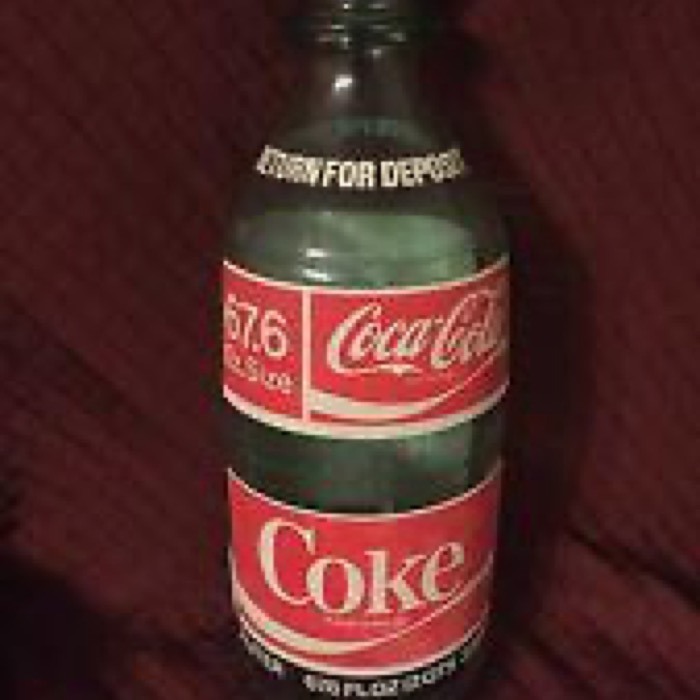} &
\qualmodplain{make the red label\\clear while keeping\\the dark brown\\curtain background} &
\qualarrowplain &
\qualframeplain{qualred}{0.135\textwidth}{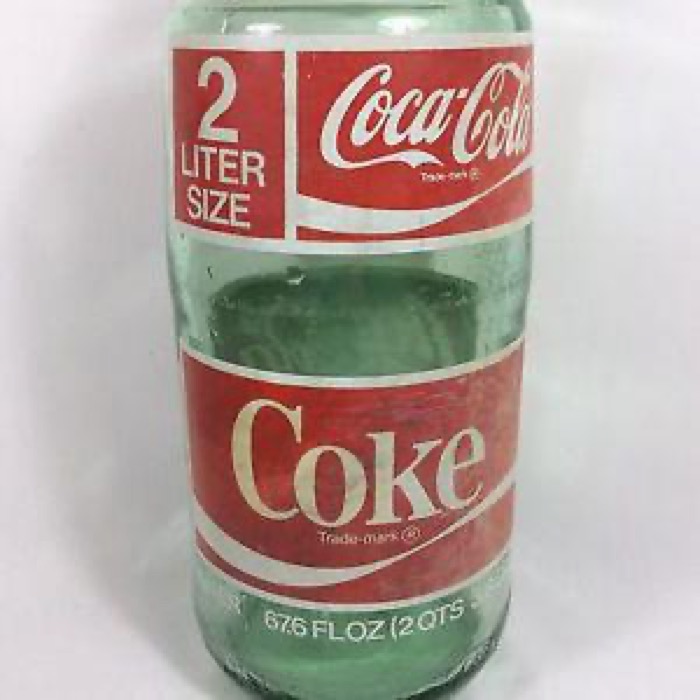} &
\qualframeplain{qualgreen}{0.135\textwidth}{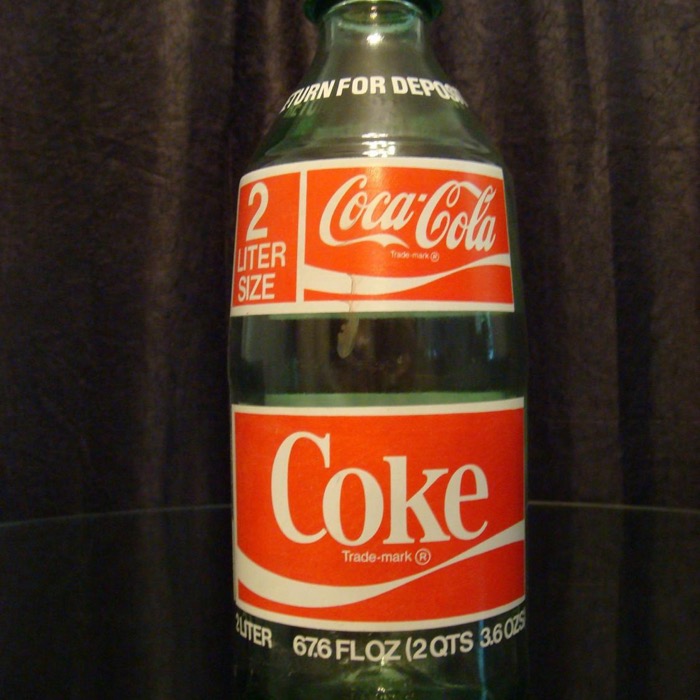}
\end{tabular}
\caption{\textbf{Qualitative comparison on CIRR.} Pic2Word can follow parts of the modification while dropping source-conditioned evidence; DeCIR better preserves reference cues while applying the requested change.}
\label{fig:main_endpoint_shortcut_cases}
\vspace{-0.6em}
\end{figure}

Figure~\ref{fig:main_endpoint_shortcut_cases} presents qualitative comparisons from CIRR that illustrate endpoint shortcut cases. Pic2Word can follow target-side cues in the modification, such as count or category, but may drop source-conditioned evidence from the reference. In contrast, DeCIR better preserves the reference cue while applying the requested change, suggesting that semantic transition learning mitigates the endpoint shortcut and better matches the CIR objective: retrieving an endpoint that is reachable from the reference under the instruction.

\section{Limitations}
DeCIR relies on an offline LLM-generated supervision pipeline. Although the LLM is not used at inference, training still depends on the quality and diversity of generated edit tuples, and investigating richer or more visually grounded instruction sources is an important future direction. This paper focuses on projection-based ZS-CIR for its lightweight deployment advantages; extending conflict-aware endpoint--transition decoupling to LLM-based CIR pipelines, stronger vision-language backbones, or fully supervised CIR settings remains to be systematically studied.
\section{Conclusion}

Projection-based ZS-CIR is fundamentally limited by weak supervision for source-conditioned semantic transitions, which often causes the model to collapse into an endpoint shortcut. DeCIR addresses this semantic transition bottleneck through conflict-aware endpoint-transition decoupling: paired supervision explicitly provides semantic transition signals, decoupled learning separates endpoint and transition alignment, and LRDM restores a single lightweight inference path. Consequently, DeCIR significantly strengthens projection-based CIR without relying on inference-time LLMs.

\clearpage
{
\small
\bibliographystyle{plain}
\bibliography{references}
}

\appendix

\section{Additional Benchmark Results}
\label{app:additional_benchmark_results}

Table~\ref{tab:genecis_main} provides the detailed GeneCIS comparison across four conditional retrieval settings. This benchmark checks whether DeCIR only improves natural-image CIR benchmarks or also transfers to conditional retrieval where the model must either focus on or change attributes and objects. DeCIR obtains the strongest average R@1 under both ViT-B/32 and ViT-L/14. While it is not uniformly best in every subcondition, the balanced average supports our claim that semantic transition learning improves source-conditioned retrieval rather than overfitting to one edit type.

\begin{table}[H]
\centering
\caption{\textbf{Quantitative comparison on the test set of GeneCIS.} We report the R@$K$ metrics across ViT-B/32 and ViT-L/14 backbones. Best results are in bold, second-best results are underlined, and our method is highlighted in grey.}
\scriptsize
\renewcommand{\arraystretch}{1.12}
\setlength{\tabcolsep}{2.4pt}
\resizebox{\textwidth}{!}{
\begin{tabular}{ll*{13}{c}}
\toprule
\multirow{2}{*}{} & \multirow{2}{*}{Method} & \multicolumn{3}{c}{Focus Attribute} & \multicolumn{3}{c}{Change Attribute} & \multicolumn{3}{c}{Focus Object} & \multicolumn{3}{c}{Change Object} & \multicolumn{1}{c}{Average} \\
\cmidrule(lr){3-5} \cmidrule(lr){6-8} \cmidrule(lr){9-11} \cmidrule(lr){12-14} \cmidrule(lr){15-15}
 &  & R@1 & R@2 & R@3 & R@1 & R@2 & R@3 & R@1 & R@2 & R@3 & R@1 & R@2 & R@3 & R@1 \\
\midrule
\multirow[c]{4}{*}{\bbcell{ViT-B/32}} & SEARLE (ICCV'23) & \underline{18.9} & 30.6 & 41.2 & 13.0 & 23.8 & 33.7 & 12.2 & 23.0 & \underline{33.3} & 13.6 & 23.8 & 33.3 & 14.4 \\
 & CIReVL (ICLR'24) & 17.9 & 29.4 & 40.4 & \underline{14.8} & \underline{25.8} & 35.8 & \underline{14.6} & \underline{24.3} & 33.3 & 16.1 & 27.8 & \underline{37.6} & 15.9 \\
 & DistillCIR (ICCV'25) & 20.0 & \underline{31.0} & \underline{42.5} & \textbf{15.2} & \textbf{26.9} & \underline{35.9} & 13.2 & 23.8 & 33.0 & \underline{16.7} & \underline{28.2} & 37.1 & \underline{16.3} \\
\rowcolor{llmrow} & \textbf{DeCIR (ours)} & \textbf{20.0} & \textbf{32.4} & \textbf{43.5} & 14.0 & 25.7 & \textbf{36.8} & \textbf{14.9} & \textbf{26.7} & \textbf{36.1} & \textbf{17.2} & \textbf{28.5} & \textbf{38.5} & \textbf{16.5} \\
\midrule
\multirow[c]{8}{*}{\bbcell{ViT-L/14}} & Pic2Word (CVPR'23) & 15.7 & 28.2 & 38.7 & 13.9 & 24.7 & 33.1 & 8.4 & 18.0 & 25.8 & 6.7 & 15.1 & 24.0 & 11.2 \\
 & SEARLE (ICCV'23) & 17.1 & 29.6 & 40.7 & 16.3 & 25.2 & 34.2 & 12.0 & 22.2 & 30.9 & 12.0 & 24.1 & 33.9 & 14.4 \\
 & LinCIR (CVPR'24) & 16.9 & 30.0 & 41.5 & 16.2 & 28.0 & 36.8 & 8.3 & 17.4 & 26.2 & 7.4 & 15.7 & 25.0 & 12.2 \\
 & Context-I2W (AAAI'24) & 17.2 & 30.5 & 41.7 & 16.4 & 28.3 & 37.1 & 8.7 & 17.9 & 26.9 & 7.7 & 16.0 & 25.4 & 12.7 \\
 & CIReVL (ICLR'24) & 19.5 & 31.8 & 42.0 & 14.4 & 26.0 & 35.2 & 12.3 & 21.8 & 30.5 & \underline{17.2} & \textbf{28.9} & \underline{37.6} & 15.9 \\
 & PrediCIR (CVPR'25) & 18.2 & 31.9 & 42.6 & \textbf{18.7} & \textbf{30.4} & 35.4 & 12.7 & 19.0 & 31.2 & 16.9 & 25.5 & 34.1 & 16.6 \\
 & DistillCIR (ICCV'25) & \underline{20.7} & \underline{32.6} & \textbf{44.2} & 16.1 & 27.6 & \underline{37.7} & \underline{13.3} & \textbf{24.4} & \underline{33.6} & \textbf{17.3} & \underline{28.8} & \textbf{37.7} & \underline{16.8} \\
\rowcolor{llmrow} & \textbf{DeCIR (ours)} & \textbf{21.1} & \textbf{32.7} & \underline{43.1} & \underline{16.4} & \underline{28.3} & \textbf{39.0} & \textbf{14.4} & \underline{23.4} & \textbf{33.8} & 16.3 & 27.2 & 36.9 & \textbf{17.1} \\
\bottomrule
\end{tabular}
}

\label{tab:genecis_main}
\end{table}
\renewcommand{\arraystretch}{1.0}
\FloatBarrier

\section{Supervision Construction and Training Details}
\label{app:supervision_training}

\label{app:llm_supervision}

This section documents how DeCIR obtains semantic transition supervision without annotated CIR triplets. Each CC3M image-caption pair is converted into a structured tuple containing a forward instruction, a modified target caption, and a reverse instruction. The target caption provides the endpoint used by endpoint alignment, while the forward and reverse instructions provide opposite semantic transition signals. The LLM is used only for offline supervision construction and is not used during inference.

We construct the supervision offline with \texttt{glm-4-flash} through the ZhipuAI batch \texttt{/v4/chat/completions} API, with the temperature set to 0.7. For each source caption, the prompt asks the LLM to choose exactly one plausible visual change, apply that change while preserving unrelated source details, and write a reverse instruction that undoes only the forward edit. We parse the batch outputs as JSON and discard malformed records or outputs missing any required fields. This format filtering leaves 667,229 clean tuples. The \texttt{brainstorming} field is used only to make generation more controlled and is not used by any training loss. Code and generated supervision will be released after acceptance.

Table~\ref{tab:llm_clean_examples} presents randomly sampled tuples generated by our pipeline. As these examples illustrate, the generated target captions successfully apply a single, fine-grained visual modification—such as altering an object's color, substituting a tool, or shifting an action—while strictly preserving all unrelated contextual details from the original source evidence. Furthermore, each forward edit is coupled with a precise reverse instruction that dictates the exact opposite semantic transition, establishing a robust, bidirectional constraint.

To ensure the quality and consistency of these generated pairs, Figure~\ref{fig:llm_prompts} details the prompt template employed for offline supervision construction. The prompt is meticulously structured to guide the LLM through a step-by-step reasoning process. By enforcing a strict JSON output contract, the prompt guarantees that the model simultaneously yields the forward edit, the modified caption, and the reverse instruction in a single pass. Crucially, it incorporates explicit constraints to ensure that only one significant visual attribute is manipulated at a time, thereby maintaining the coherence and plausibility of the overall scene.

\begin{table}[!htbp]
\centering
\caption{\textbf{Randomly sampled tuples.} Each example applies one visual change, preserves unrelated source evidence in the target caption, and provides a reverse instruction for the opposite transition.}
\small
\renewcommand{\arraystretch}{1.18}
\setlength{\tabcolsep}{4pt}
\begin{tabular}{>{\raggedright\arraybackslash}p{0.24\linewidth} >{\raggedright\arraybackslash}p{0.48\linewidth} >{\raggedright\arraybackslash}p{0.21\linewidth}}
\toprule
\rowcolor{black!8}
Source evidence & Forward tuple & Reverse instruction \\
\midrule
open red umbrella casting a shadow &
\textbf{Forward:} Change the color of the umbrella from red to blue.\newline
\textbf{Target:} An open blue umbrella with a shadow. &
Change the color of the umbrella from blue to red. \\
\rowcolor{llmrow}
worker clearing snow off the sidewalk with a broom &
\textbf{Forward:} Replace the broom with a shovel for clearing snow.\newline
\textbf{Target:} A worker clears snow off the sidewalk with a shovel. &
Replace the shovel with a broom for clearing snow. \\
dog standing on a balcony and looking down &
\textbf{Forward:} Change the dog's posture from standing to sitting.\newline
\textbf{Target:} A dog sits on a balcony looking down. &
Change the dog's posture from sitting to standing. \\
\rowcolor{llmrow}
young man struggling to row through a river &
\textbf{Forward:} Change the river to a lake.\newline
\textbf{Target:} A young man struggles to row through a lake. &
Change the lake to a river. \\
statue of a builder at a shrine, originally stone &
\textbf{Forward:} Change the material of the statue from stone to gold.\newline
\textbf{Target:} A gold statue of a builder at the shrine. &
Change the material of the statue to stone. \\
\bottomrule
\end{tabular}

\label{tab:llm_clean_examples}
\end{table}

\begin{figure}[!htbp]
\centering
\begingroup
\setlength{\fboxsep}{10pt}
\setlength{\fboxrule}{0.6pt}
\fcolorbox{gray!35}{white}{\begin{minipage}{0.90\linewidth}
\footnotesize
\setlength{\parindent}{0pt}
\setlength{\parskip}{0.45em}
\textbf{You are helping to create a multimodal dataset for Composed Image Retrieval (CIR). Given one source image caption, create one plausible visual edit from source to target, a modified target caption, and the reverse instruction that would transform the modified caption back to the source.}

\textbf{Task}\\
1. Input: A source image caption will be provided.\\
2. Brainstorming: Identify the key source details in the source caption (objects, actions, setting) and propose one significant, plausible visual change.\\
3. \texttt{instruction}: Write the forward edit from source to target.\\
4. \texttt{modified\_caption}: Apply the edit while preserving unrelated source details.\\
5. \texttt{reverse\_instruction}: Write the reverse edit from \texttt{modified\_caption} back to the source.

\textbf{Output Requirements}\\
Return valid JSON only with exactly these keys:\\
1. \texttt{"brainstorming"} -- Briefly explain the source details and the proposed change.\\
2. \texttt{"instruction"} -- A short statement of the exact source-to-target change.\\
3. \texttt{"modified\_caption"} -- The new target caption after applying the edit.\\
4. \texttt{"reverse\_instruction"} -- A short instruction that undoes only the forward edit.

\textbf{Important}\\
1. Make exactly one significant visual change, such as changing an object color, location, action, material, count, or category.\\
2. Keep unrelated details unchanged in \texttt{modified\_caption}.\\
3. The \texttt{reverse\_instruction} must undo only the forward edit.\\
4. The instruction and modified caption should be coherent and plausible.

\textbf{Input: \{source\_caption\}}
\end{minipage}}
\endgroup
\caption{\textbf{Prompt template used for offline supervision construction.} We prompt the LLM to produce the forward edit, edited caption, and reverse instruction in one structured output.}
\label{fig:llm_prompts}
\end{figure}

\FloatBarrier

\label{app:hyperparameters}

\begin{table}[H]
\centering
\caption{\textbf{Main DeCIR training hyperparameters.} Common optimization settings follow the DistillCIR-style setup where applicable, while DeCIR-specific values such as the LRDM merge weight and source-anchor image weight are shown explicitly.}
\footnotesize
\renewcommand{\arraystretch}{1.16}
\setlength{\tabcolsep}{4pt}
\begin{tabular}{>{\raggedright\arraybackslash}p{0.27\linewidth} >{\raggedright\arraybackslash}p{0.18\linewidth}| >{\raggedright\arraybackslash}p{0.27\linewidth} >{\raggedright\arraybackslash}p{0.15\linewidth}}
\toprule
Configuration & Value & Configuration & Value \\
\midrule
Visual tower & CLIP ViT & Optimizer & AdamW \\
Text tower & CLIP text encoder & Learning rate & $2\times10^{-5}$ \\
Clean generated tuples & 667,229 & Weight decay & 0.1 \\
LoRA rank & 64 & Warmup & 200 steps \\
LoRA alpha & 16 & LR scheduler & Cosine \\
Precision & FP16 & LRDM merge weight $\alpha$ & 0.50 \\
Training epochs & 1 & Source anchor image weight $\omega$ & 0.25 \\
\bottomrule
\end{tabular}

\label{tab:hyperparameters}
\end{table}

As shown in Table~\ref{tab:hyperparameters}, we list the main training configuration used for DeCIR. We follow the DistillCIR-style training setup where applicable to keep the comparison focused on the proposed endpoint--transition decoupling. DeCIR-specific choices, including the number of clean generated tuples, LRDM merge weight $\alpha$, and source-anchor image weight $\omega$, are reported explicitly.

Algorithm~\ref{alg:decir} summarizes the training procedure of DeCIR and the LRDM merge. The key implementation detail is update ownership: the endpoint alignment branch updates the endpoint-aligned retrieval pathway and the shared low-rank bases, while the transition alignment branch computes retrieval-space anchors, stops gradients through these anchors, and updates only the transition coefficients $A_{\mathrm{trans}}$. This enables training-time decoupling while keeping the two branches in a shared coefficient space for LRDM.

\begin{algorithm}[H]
\caption{\textbf{DeCIR training and LRDM merge.}}
\label{alg:decir}
\begin{algorithmic}[1]
\Require Batch $\mathcal{B}=\{(I_{\mathrm{ref}},c_{\mathrm{src}},t_{\mathrm{fwd}},c_{\mathrm{tgt}},t_{\mathrm{rev}})\}$; shared bases $\{B^\ell\}$; branch coefficients $\{A_{\mathrm{end}}^\ell,A_{\mathrm{trans}}^\ell\}$; merge weight $\alpha$
\Statex \textbf{Training phase}
\For{each mini-batch $\mathcal{B}$}
  \State Build composed queries with $E_T^{\mathrm{end}}$, $E_V$, and $f_{\phi}$ using $(I_{\mathrm{ref}},t_{\mathrm{fwd}})$.
  \State Compute $\mathcal{L}_{\mathrm{end}}$ against target caption embedding $f_{\mathrm{tgt}}^{\mathrm{end}}$.
  \State Update $\{A_{\mathrm{end}}^\ell\}$, shared $\{B^\ell\}$, visual LoRA parameters, $f_{\phi}$, and $\tau$.
  \State Compute $f_{\mathrm{src}}$, $f_{\mathrm{tgt}}^{\mathrm{trans}}$, and $\delta$ and detach.
  \State Encode $t_{\mathrm{fwd}}$ and $t_{\mathrm{rev}}$ with $E_T^{\mathrm{trans}}$ and compute $\mathcal{L}_{\mathrm{trans}}$.
  \State Update only $\{A_{\mathrm{trans}}^\ell\}$.
\EndFor
\Statex \textbf{Merge phase}
\For{each adapted text layer $\ell$}
  \State $A_{\mathrm{merge}}^\ell \gets (1-\alpha)A_{\mathrm{end}}^\ell+\alpha A_{\mathrm{trans}}^\ell$, \quad $B_{\mathrm{merge}}^\ell\gets B^\ell$
  \State $W'^\ell \gets W^\ell+B_{\mathrm{merge}}^\ell A_{\mathrm{merge}}^\ell$
\EndFor
\State \Return final retrieval model $\{E_V,f_{\phi},E_T(W')\}$
\end{algorithmic}
\end{algorithm}
\FloatBarrier

\section{Additional Ablations}
\label{app:additional_merge}
\label{app:additional_ablations}

Table~\ref{tab:merge_strategy_ablation} tests whether DeCIR's improvement can be obtained by applying off-the-shelf model merging rules to the same trained branches, including Task Arithmetic~\citep{ilharco2023editingmodelstaskarithmetic}, TIES~\citep{yadav2023ties}, DARE~\citep{yu2024language}, Model Breadcrumbs~\citep{modelbreadcrumbs}, and RobustMerge~\citep{zeng2025parameter}. We keep the retrieval and transition branches fixed and change only the merge rule. LRDM performs best across all three validation metrics, showing that the gain is not simply due to arbitrary parameter averaging. Instead, it supports the asymmetric design of LRDM: the endpoint alignment branch preserves the deployable endpoint-aligned backbone, while the transition branch contributes semantic transition coefficients in the shared low-rank basis.

\begin{figure}[!htbp]
\vspace{-1.8em} % 【往上一点】通过负的垂直间距把图表往上提（数值可根据需要调整，如 -1em, -2em）
\centering
\makebox[\textwidth][c]{%
    \hspace{-2.2em}% 【往左一点】在这里输入负值，强行把图片往左拽（数值可微调，如 -1em, -3em）
    \includegraphics[width=0.50\linewidth]{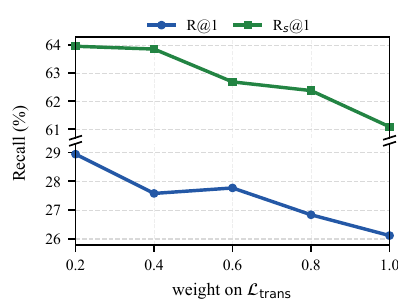}%
}
\vspace{-9pt}
\caption{\textbf{Shared-adapter transition-loss weight sweep.} We sweep $\lambda_{\mathrm{trans}}$ in $\mathcal{L}_{\mathrm{end}}+\lambda_{\mathrm{trans}}\mathcal{L}_{\mathrm{trans}}$ on CIRR validation. The sweep supports that joint underperformance is not simply due to an untuned transition-loss scale.}
\label{fig:joint_transition_loss_sweep_app}
\end{figure}

Figure~\ref{fig:joint_transition_loss_sweep_app} tests whether the failure of shared-adapter joint training is merely a loss-weighting issue. We optimize the same shared adapter with $\mathcal{L}_{\mathrm{end}}+\lambda_{\mathrm{trans}}\mathcal{L}_{\mathrm{trans}}$ and sweep $\lambda_{\mathrm{trans}}$ on the CIRR validation set. If the problem were only a loss-scale imbalance, some weights should approach the decoupled result in Table~\ref{tab:core_ablation}. Instead, increasing the transition weight generally degrades retrieval, and the shared-adapter objective does not close the gap to the decoupled learning. This supports our claim that semantic transition supervision should be integrated through conflict-aware decoupling rather than treated as a standard auxiliary loss.

\begin{table}[H]
\centering
\caption{\textbf{Merge strategy ablation on validation sets.} The trained retrieval and transition branches are fixed; only the merge rule changes. All merge weights are set to 0.5. LRDM outperforms generic merge rules, supporting shared-basis coefficient merging rather than generic weight-space merging.}
\scriptsize
\renewcommand{\arraystretch}{1.10}
\setlength{\tabcolsep}{6pt}
\begin{tabular}{lccc}
\toprule
Merge strategy & CIRR R$_s$@1 & CIRCO mAP@50 & GeneCIS Avg. R@1 \\
\midrule
Task Arithmetic~\citep{ilharco2023editingmodelstaskarithmetic} & 65.22 & 27.07 & 16.57 \\
TIES~\citep{yadav2023ties} & 63.29 & 26.70 & 16.49 \\
TIES (frequency)~\citep{yadav2023ties} & 63.45 & 26.86 & 16.31 \\
DARE (linear)~\citep{yu2024language} & 64.96 & 27.05 & 16.55 \\
DARE+TIES~\citep{yu2024language,yadav2023ties} & 63.38 & 26.75 & 16.45 \\
Model Breadcrumbs~\citep{modelbreadcrumbs} & 64.41 & 27.14 & 16.55 \\
RobustMerge~\citep{zeng2025parameter} & 65.20 & 27.06 & 16.60 \\
LRDM & 66.08 & 27.26 & 16.86 \\
\bottomrule
\end{tabular}

\label{tab:merge_strategy_ablation}
\end{table}

Table~\ref{tab:adapter_scope_ablation} isolates the trainable component scope while keeping LRDM fixed. Text-adapter-only training underperforms the setting that also adapts the visual encoder and trains the full inversion network. This suggests that source-conditioned composed retrieval benefits from adapting the pathway that exposes reference-image evidence to the text encoder, not only the text adapter that receives the modification. We therefore use LoRA on both CLIP towers and train the full inversion network.

\begin{figure}[!htbp]
\centering
\includegraphics[width=0.98\linewidth]{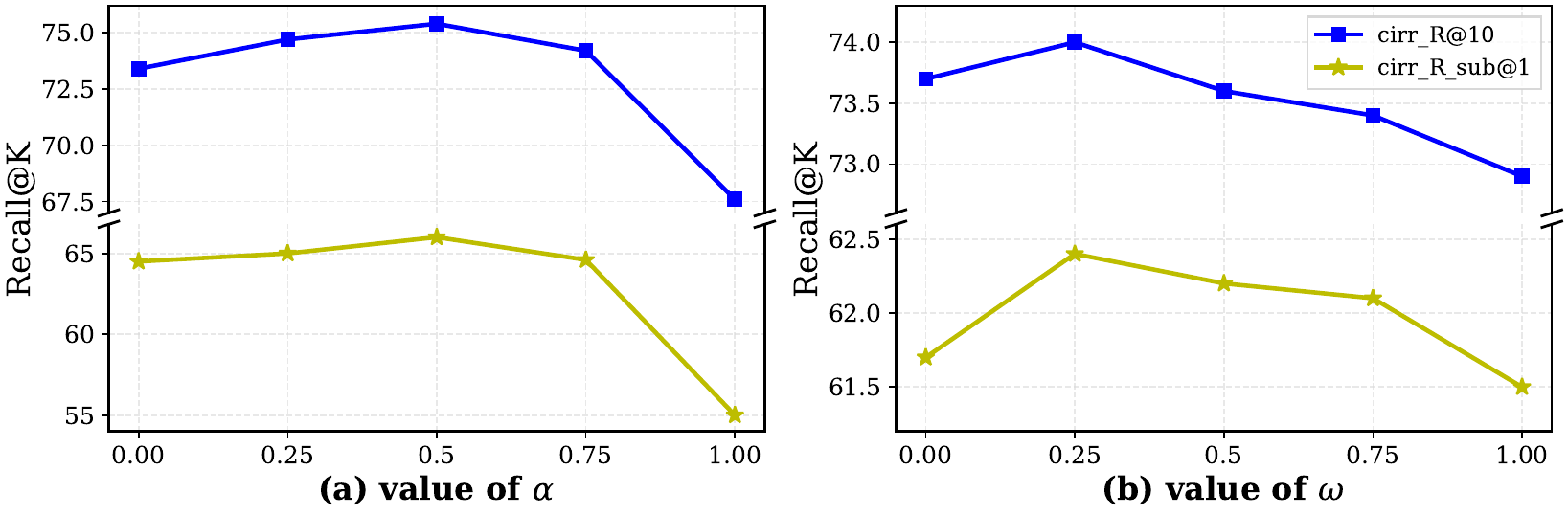}
\caption{\textbf{Parameter sensitivity analysis on CIRR validation set.} The $\alpha$ sweep shows that transition coefficients work best when injected into an endpoint alignment branch rather than used alone. The $\omega$ sweep studies how much image-conditioned source evidence should enter the transition anchor.}
\label{fig:merge_weight_sensitivity}
\end{figure}

Figure~\ref{fig:merge_weight_sensitivity} studies two DeCIR-specific weights. The LRDM weight $\alpha$ controls how much transition-branch coefficient information is injected into the endpoint alignment branch. The endpoints of the sweep are informative: $\alpha=0$ reduces to retrieval-only inference, while $\alpha=1$ uses pure transition coefficients. The best performance occurs at an intermediate value, supporting our view that semantic transition learning is complementary residual knowledge rather than a standalone low-rank adapter. The source-anchor weight $\omega$ controls how much image-conditioned evidence enters the source anchor. The sensitivity curve shows that incorporating visual source evidence is useful, while an overly image-dominated anchor is not necessary.

\begin{table}[H]
\centering
\caption{\textbf{Trainable component ablation on validation sets.} The merge rule is fixed to LRDM; only the trainable component scope changes. This implementation ablation justifies the final adaptation scope used by DeCIR.}
\scriptsize
\renewcommand{\arraystretch}{1.10}
\setlength{\tabcolsep}{6pt}
\begin{tabular}{lccc}
\toprule
Trainable components & CIRR R$_s$@1 & CIRCO mAP@50 & GeneCIS Avg. R@1 \\
\midrule
LoRA on $E_T$ & 60.11 & 23.10 & 14.88 \\
LoRA on $E_T$, full on $f_{\phi}$ & 60.32 & 23.97 & 14.94 \\
LoRA on $(E_V,E_T)$, full on $f_{\phi}$ & 66.08 & 27.26 & 16.86 \\
\bottomrule
\end{tabular}

\label{tab:adapter_scope_ablation}
\end{table}
\FloatBarrier

\section{Deployment Profile}
\label{app:deployment_profile}

\begin{table}[H]
\caption{\textbf{Deployment profile in the ViT-L/14 setting.} DeCIR uses LLM supervision only offline during training. After LRDM, it deploys a single merged retrieval model with no LLM or second text branch at inference.}
\centering
\footnotesize
\renewcommand{\arraystretch}{1.12}
\setlength{\tabcolsep}{7pt}
\begin{tabular}{lccc}
\toprule
Method & LLM at inference & Trainable params & Inference params \\
\midrule
Context-I2W (AAAI'24) & $\times$ & 65.3M & 493M \\
CIReVL (ICLR'24) & \checkmark & -- & 12.5B \\
MCL (ICML'24) & \checkmark & 25.8M & 7.4B \\
DistillCIR (ICCV'25) & $\times$ & 42M & 467M \\
WISER (CVPR'26) & \checkmark & -- & $\geq$33.4B \\
\textbf{DeCIR (ours)} & $\times$ & 50M & 467M \\
\bottomrule
\end{tabular}

\label{tab:deployment_profile}
\end{table}

Table~\ref{tab:deployment_profile} compares the deployment footprint of DeCIR with representative ZS-CIR systems, including Context-I2W~\citep{tang2024context}, CIReVL~\citep{karthik2024vision}, MCL~\citep{mcl}, DistillCIR~\citep{zhong2025zero}, and WISER~\citep{wang2026wiser}. DeCIR uses LLM-generated supervision only offline during training; after LRDM, inference uses one visual encoder, one inversion network, and one merged text adapter. Thus, its active inference parameter count remains close to projection-based methods such as Context-I2W and DistillCIR, while avoiding the online LLM or reasoning modules used by heavier systems. Combined with the benchmark results in the main text, this supports the claim that DeCIR improves the projection-based accuracy--efficiency trade-off rather than trading accuracy for an inference-time reasoning loop.

\FloatBarrier

\section{Qualitative Analysis}
\label{app:qualitative}

\label{app:hard_distractors}

\begin{figure}[H]
\centering
\setlength{\tabcolsep}{3.8pt}
\renewcommand{\arraystretch}{1.0}
\begin{tabular}{@{}>{\centering\arraybackslash}m{0.18\textwidth}
                  >{\centering\arraybackslash}m{0.18\textwidth}
                  >{\centering\arraybackslash}m{0.035\textwidth}
                  >{\centering\arraybackslash}m{0.18\textwidth}
                  >{\centering\arraybackslash}m{0.18\textwidth}@{}}
\multicolumn{1}{c}{\qualhead{Reference}} &
\multicolumn{1}{c}{\qualhead{Modification}} & &
\multicolumn{1}{c}{\qualhead{Pic2Word}} &
\multicolumn{1}{c}{\qualhead{DeCIR}} \\
\qualframeplain{qualgray}{0.17\textwidth}{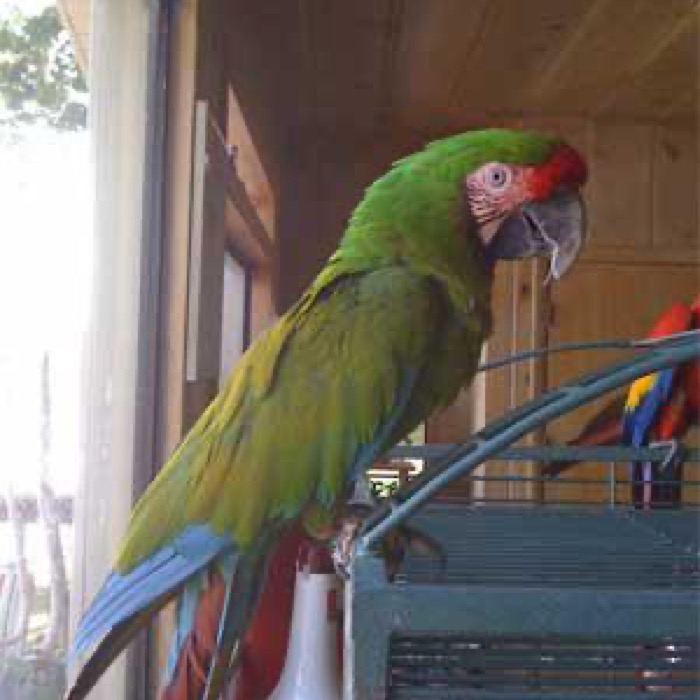} &
\qualmodplain{make it\\two birds} &
\qualarrowplain &
\qualframeplain{qualred}{0.17\textwidth}{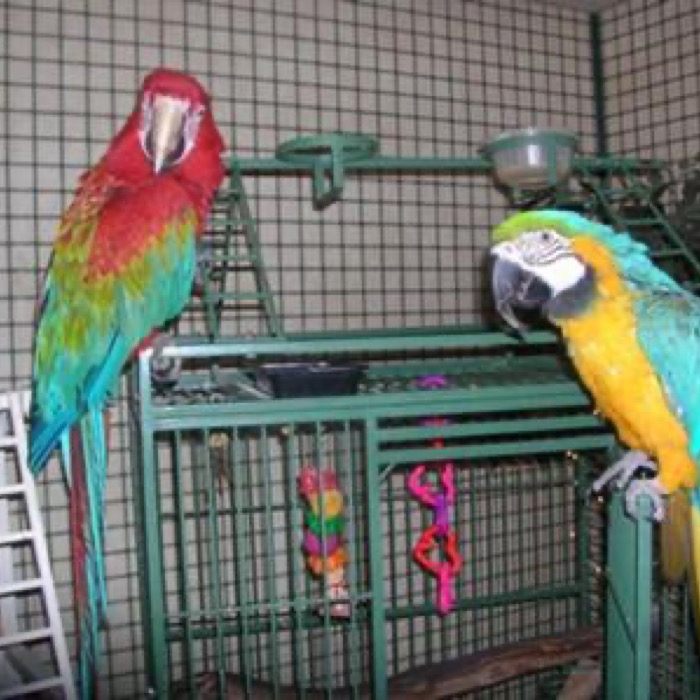} &
\qualframeplain{qualgreen}{0.17\textwidth}{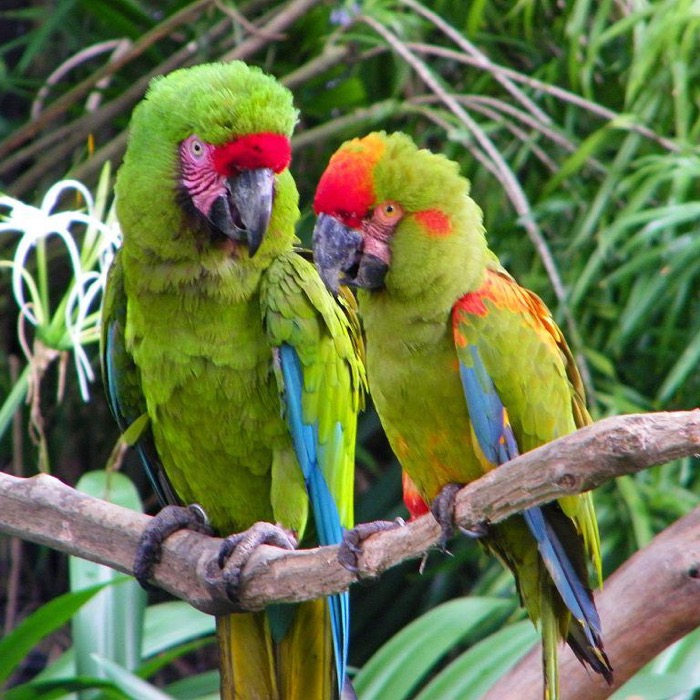} \\
\addlinespace[0.55em]
\qualframeplain{qualgray}{0.17\textwidth}{figures/fig8_hard_cirr_q1439_ref.jpg} &
\qualmodplain{make the red label\\clear while keeping\\the dark brown\\curtain background} &
\qualarrowplain &
\qualframeplain{qualred}{0.17\textwidth}{figures/fig8_hard_cirr_q1439_ret.jpg} &
\qualframeplain{qualgreen}{0.17\textwidth}{figures/fig8_hard_cirr_q1439_decir.jpg} \\
\addlinespace[0.55em]
\qualframeplain{qualgray}{0.17\textwidth}{figures/fig8_hard_cirr_q1204_ref.jpg} &
\qualmodplain{change to a herd\\of three wild\\mountain goats\\with a tighter rocky\\mountain view} &
\qualarrowplain &
\qualframeplain{qualred}{0.17\textwidth}{figures/fig8_hard_cirr_q1204_ret.jpg} &
\qualframeplain{qualgreen}{0.17\textwidth}{figures/fig8_hard_cirr_q1204_decir.jpg}
\end{tabular}
\caption{\textbf{Qualitative comparison on CIRR.} The Pic2Word can follow parts of the modification but drop source-conditioned evidence. DeCIR retrieves images that satisfy the modification while preserving the reference constraint.}
\label{fig:hard_distractor_cases}
\end{figure}

\begin{figure}[H]
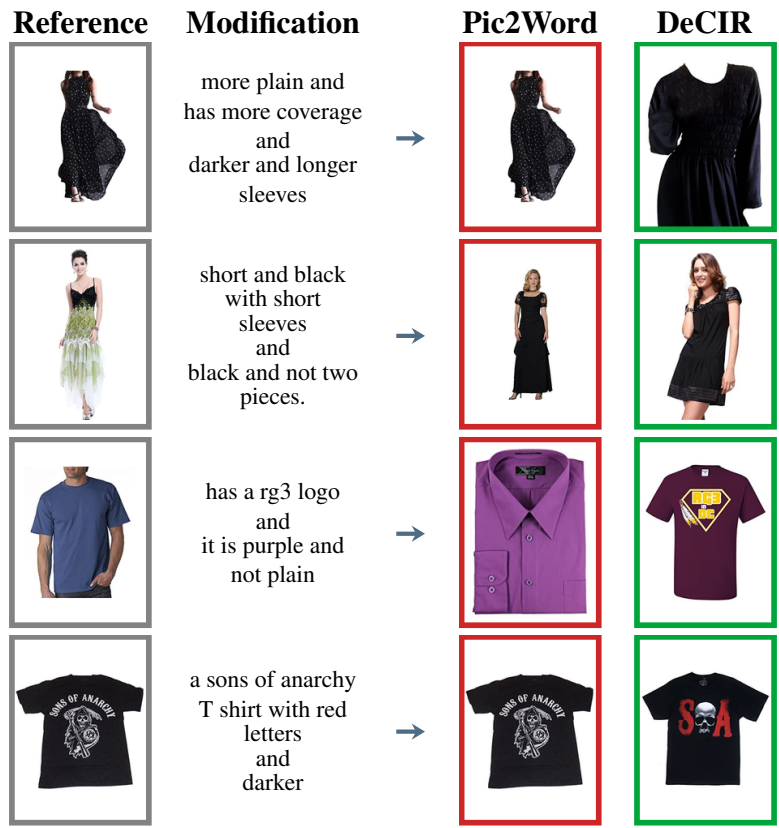

\centering
\setlength{\tabcolsep}{6pt}
\renewcommand{\arraystretch}{1.0}
\begin{tabular}{@{}ccccc@{}}
\qualhead{Reference} & \qualhead{Modification} & & \qualhead{Pic2Word} & \qualhead{DeCIR} \\
\fashioncrop{26}{780}{1074}{77} &
\qualmod{more plain and\\has more coverage\\and\\darker and longer\\sleeves} &
\qualarrow &
\fashioncrop{504}{780}{596}{77} &
\fashioncrop{1074}{780}{26}{77} \\
\addlinespace[0.25em]
\fashioncrop{26}{527}{1074}{329} &
\qualmod{short and black\\with short\\sleeves\\and\\black and not two\\pieces.} &
\qualarrow &
\fashioncrop{504}{527}{596}{329} &
\fashioncrop{1074}{527}{26}{329} \\
\addlinespace[0.25em]
\fashioncrop{26}{275}{1074}{581} &
\qualmod{has a rg3 logo\\and\\it is purple and\\not plain} &
\qualarrow &
\fashioncrop{504}{275}{596}{581} &
\fashioncrop{1074}{275}{26}{581} \\
\addlinespace[0.25em]
\fashioncrop{26}{23}{1074}{833} &
\qualmod{a sons of anarchy\\T shirt with red\\letters\\and\\darker} &
\qualarrow &
\fashioncrop{504}{23}{596}{833} &
\fashioncrop{1074}{23}{26}{833}
\end{tabular}
\caption{\textbf{Qualitative comparison on FashionIQ.} The examples cover fashion modifications involving coverage, sleeve length, color, logos, and style. DeCIR improves edit sensitivity while preserving source garment evidence. Green boxes indicate correct top-1 retrievals, and red boxes indicate incorrect top-1 retrievals.}
\label{fig:fashioniq_qualitative}
\end{figure}

Figure~\ref{fig:hard_distractor_cases} presents qualitative comparisons from CIRR. The examples cover natural-image modifications involving object count, text clarity, and scene composition. The Pic2Word often captures part of the target-side change but can lose source-conditioned evidence from the reference, whereas DeCIR better preserves the source constraint while applying the requested modification. Similarly, Figure~\ref{fig:fashioniq_qualitative} extends this analysis to the FashionIQ dataset, highlighting fine-grained fashion edits such as changes in coverage, sleeve length, color, logos, and style. In these cases, DeCIR demonstrates improved edit sensitivity, successfully retrieving the correct top-1 targets (indicated by green boxes) by applying the desired changes while accurately preserving the essential visual evidence of the source garment.

\clearpage

\end{document}